\documentclass[runningheads]{llncs}

\usepackage{eccv}

\usepackage{eccvabbrv}

\usepackage{graphicx}
\usepackage{booktabs}

\usepackage[accsupp]{axessibility}  

\usepackage[hidelinks]{hyperref}

\usepackage{orcidlink}

\begin{document}

\title{SICAGE: Speaker-Independent Culture-Aware Gesture Generation using
TED4C-L Dataset}

\titlerunning{SICAGE: Speaker-Independent Culture-Aware Gesture Generation}

\author{Ariel Gjaci\inst{1}\orcidlink{0000-0001-9794-5200}\textsuperscript{*} \and
Antonio Sgorbissa\inst{2}\orcidlink{0000-0001-7789-4311} \and
Vittorio Murino\inst{1}\orcidlink{0000-0002-8645-2328}}

\authorrunning{A.~Gjaci et al.}

\institute{Italian Institute of Technology, Genoa, Italy\\
\textsuperscript{*}\email{ariel.gjaci@iit.it} \and
University of Genoa, Genoa, Italy 
}

\maketitle

\begin{abstract}
Recent co-speech gesture generation methods often overlook cultural differences, limiting their effectiveness in human–agent interaction. Moreover, culture-conditioned models are rarely evaluated under speaker-disjoint splits, so apparent “cultural” behavior may be confounded with speaker-specific gesturing style. We introduce SICAGE, a modular framework for culture-aware co-speech gesture generation that conditions motion synthesis models on speaker-independent cultural representations. SICAGE learns these representations from audio and text by treating each speaker as a separate domain while imposing invariance across speakers. This encourages representations to remain culture-discriminative while reducing dependence on speaker identity. The resulting cultural embeddings condition a multimodal generator to produce culturally appropriate gestures. We instantiate this idea with two domain generalization approaches: adversarial learning and Fishr regularization. We further introduce ALaDiT, a real-time diffusion-based gesture generator designed to efficiently incorporate the learned cultural embeddings.
To validate our method, we built TED4C-L, a 106-hour multimodal dataset of 764 TED speakers from four cultural groups. Experiments show that SICAGE improves motion realism, diversity, beat synchronization, semantic relevance, and cultural consistency. 
\end{abstract}

\begin{figure} 
\centering
\includegraphics[width=1\columnwidth]{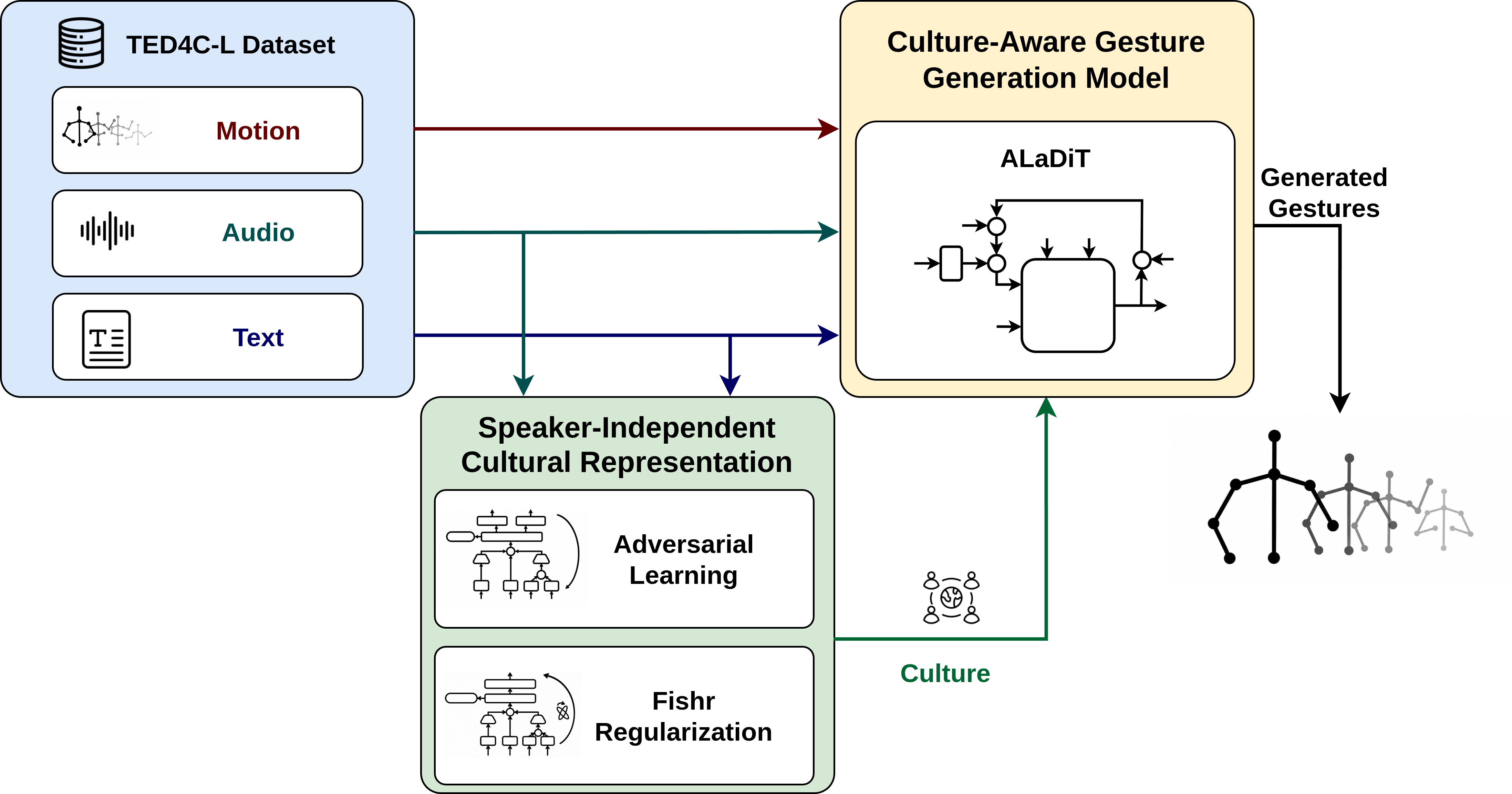}
\caption{Overview of our implementation of SICAGE. We extract text, audio, and motion features from TED4C-L (blue), regularize text/audio into speaker‑independent cultural embeddings via Fishr or adversarial training (green), then feed these embeddings together with other raw inputs into ALaDiT (yellow) to generate real‑time, culture‑aware gestures. All components are modular and replaceable.}
\label{fig:SICAGE-framework}
\end{figure}

\section{Introduction}
\label{sec:intro}

Human communication is inherently multimodal, combining speech with co-speech gestures that reinforce verbal messages and convey abstract concepts and emotions. The importance of gestures is well documented in human–human~\cite{1,2,3,56}, human–agent~\cite{4,5,7-extra, 57}, and social robotics interactions~\cite{6,7}. A critical yet underexplored dimension in gesture generation is culture, as cultural norms shape gesture performance and interpretation~\cite{8,9,10}. 
Defining cultural boundaries is itself nontrivial, especially in dynamic and multiethnic societies~\cite{12}. As a result, ``culture-aware'' modeling must clearly specify what aspects of culture are being measured and how evaluation is conducted; otherwise, apparent cultural effects may simply reflect confounding factors.

Existing culture-aware gesture generation methods~\cite{31,32,11} rely on culturally annotated datasets but often do not rigorously test whether learned cultural patterns transfer across speakers. In particular, evaluations frequently use speaker-dependent protocols in which the same speakers appear in both training and test splits. In such settings, models may appear to capture culture while relying on speaker-specific cues that reflect individual style rather than group-level patterns. A robust evaluation of cultural generalization therefore requires both large-scale data and explicitly speaker-disjoint splits, so that culture must be inferred from patterns shared across speakers rather than memorized from individuals.

In this paper, we address these limitations by proposing \textbf{SICAGE}\footnote{
Project page, dataset, and source code:
\url{https://arielgjaci.com/sicage}
} (Speaker-Independent Culture-Aware Gesture gEneration), a modular framework for synthesizing culture-aware gestures while explicitly reducing dependence on speaker identity. Within SICAGE, we adopt the view of culture as a learned social and individual construct affecting both behavior and interpretation~\cite{12}, and operationally model it as speaker-disjoint, group-level communication patterns shared within cultural groups. Since these patterns are expressed through entangled linguistic, prosodic, semantic, and other behavioral cues~\cite{DG-m,55}, our goal is not to causally disentangle culture from these modalities. Instead, we learn conditioning embeddings that are discriminative of cultural groups while reducing their dependence on individual speaker style. We therefore cast cultural representation learning as a domain-generalization problem, treating each speaker as a domain and using speaker-invariant learning objectives to encourage embeddings that generalize to unseen speakers from the same cultural group.

To support robust training and evaluation under this objective, we introduce \textbf{TED4C-L}, a large-scale, multimodal, and speaker-balanced dataset with 106 hours of TED talks from 764 speakers across four regions: India (TED Talks in Hindi), Italy, Turkey, and Japan. We select these groups based on abundant native-language TED talks and clear geographic separation to minimize cultural overlap. Unlike English, these languages closely reflect geographic and cultural identity, enhancing annotation reliability. To assess whether cultural patterns are detectable from gesture motion alone, we train a motion-based classifier to predict cultural labels (see supplementary for details). Under speaker-disjoint evaluation, the classifier achieves approximately 45\% balanced accuracy on unseen speakers, compared to a 25\% random baseline for four classes. This result indicates that gesture motion contains culture-related signals that generalize across speakers, while also highlighting the substantial intra-cultural variability that makes the task challenging.

Within SICAGE, we learn speaker-invariant cultural representations using two domain generalization strategies that treat each speaker as a domain: Fishr regularization~\cite{13} and adversarial learning~\cite{49}, which has proven effective in related multimodal settings~\cite{DG-m}. These objectives encourage representations that remain predictive of culture while reducing sensitivity to speaker identity. 

For gesture synthesis, we adopt a diffusion-based motion generator~\cite{14}, ALaDiT, conditioned on audio, text, and the learned cultural embeddings. This design enables gestures that remain synchronized with speech while reflecting culture-dependent motion patterns. Figure~\ref{fig:SICAGE-framework} provides an overview of the SICAGE framework. Our experiments show that enforcing speaker-invariance improves motion quality, diversity, cultural consistency, and alignment with rhythmic and contextual features extracted from text and culture. In addition, our diffusion-based generator outperforms recent baselines on standard gesture generation metrics. 

Overall, the contributions of this work are: (i) \textbf{SICAGE}, a modular framework for speaker-independent, culture-aware co-speech gesture generation; (ii) \textbf{TED4C-L}, a large-scale, multimodal dataset explicitly designed for cultural generalization with disjoint-speaker splits; (iii) the introduction of \textbf{domain generalization} methods, instantiated with \textbf{Fishr} and \textbf{adversarial learning}, to learn cultural cues that generalize across speakers; and (iv) ALaDiT, a diffusion-based gesture generator conditioned on these cues to synthesize high-quality culture-aware motion.

\section{Related Work}
\label{sec:state-of-the-art}

\subsection{Rule-based methods}

Early gesture generation relied on rule-based systems mapping linguistic or prosodic cues to predefined gestures. BEAT~\cite{15}, for example, uses linguistic and contextual annotations to animate gestures via a knowledge-based engine. Similarly, other approaches combine part-of-speech tagging and speech timing with grammar rules to sample gesture trajectories \cite{16}. More recent works \cite{17} learn mappings from video but suffer scalability and memory issues. Although these methods produce smooth motion, they rely on predefined gesture units, and to our knowledge, only one study \cite{18} explicitly considers cultural variations.

\subsection{Data-driven methods}
With larger datasets, gesture synthesis has transitioned to data-driven models, favoring scalability over interpretability. Early probabilistic models, such as MDP controllers~\cite{19} and parameter-based selection~\cite{20}, infer gestures from prosodic features. LLM-based approaches use language models for intent extraction to generate gestures~\cite{21,22}, yet remain limited by gesture set size. End-to-end generative methods (GANs~\cite{23}, attention-based~\cite{24}, diffusion models \cite{25}) produce novel, diverse motions, with diffusion-transformer models achieving state-of-the-art quality due to tight text/audio alignment~\cite{26,27,28,29,30}. However, existing diffusion models lack explicit cultural conditioning. Some works implicitly embed cultural cues via attention~\cite{31} or culture-specific GANs~\cite{32}, but provide limited quality and generalization to unseen speakers. Our approach conditions diffusion models on dedicated, speaker-independent cultural embeddings for robust cross-cultural synthesis.

\subsection{Domain Generalization}
Domain Generalization (DG) aims to generalize to unseen domains by training on related ones~\cite{DG-survey}. Common strategies include data augmentation~\cite{DG-d1,DG-d2}, representation learning~\cite{49,DG-r1}, and advanced training schemes~\cite{13,DG-l1}. Following~\cite{DG-m}, in this work we treat each speaker in the dataset as a domain and condition gesture synthesis on cultural embeddings. We utilize adversarial learning~\cite{49}, shown to be effective in similar tasks \cite{DG-m}, and Fishr~\cite{13}, known for efficiently matching domain gradient variances. Fishr’s scalability and performance make it ideal for embedding cultural representations across numerous domains. 

\subsection{Datasets}
High-quality multimodal data is critical for gesture synthesis. Motion-capture datasets such as CMU Panoptic~\cite{33} and Talking With Hands 16.2M~\cite{34} provide accurate 3D keypoints but lack diversity; only LISI-HHI~\cite{35} provides cultural annotations, but at a small scale. TED-Talk datasets~\cite{36,51}, using OpenPose-extracted skeletons~\cite{38}, GloVe text embeddings~\cite{39}, and audio features, offer similar size and greater diversity (97 hours, 1,700+ speakers) but omit cultural labels. MCGD~\cite{31} annotates culture for 263 speakers, but it is not public and covers only $\sim$20 speakers per culture. We introduce TED4C-L, comprising approximately 190 speakers per culture across four cultures (106 hours total), explicitly designed for cultural generalization and gesture generation tasks.

\section{Methodology}
\label{sec:met}

We introduce SICAGE, a modular framework for culture-aware co-speech gesture generation comprising: (i) a culturally diverse dataset (TED4C-L in our implementation), (ii) a model for learning speaker-independent cultural representations (via Fishr regularization or adversarial learning in our implementation), and (iii) a motion generator conditioned on culture and other features (ALaDiT in our implementation). Each step can be implemented in various ways; our implementation is detailed in the following sections.

\begin{figure}[t]
\centering
\includegraphics[width = .96\textwidth]{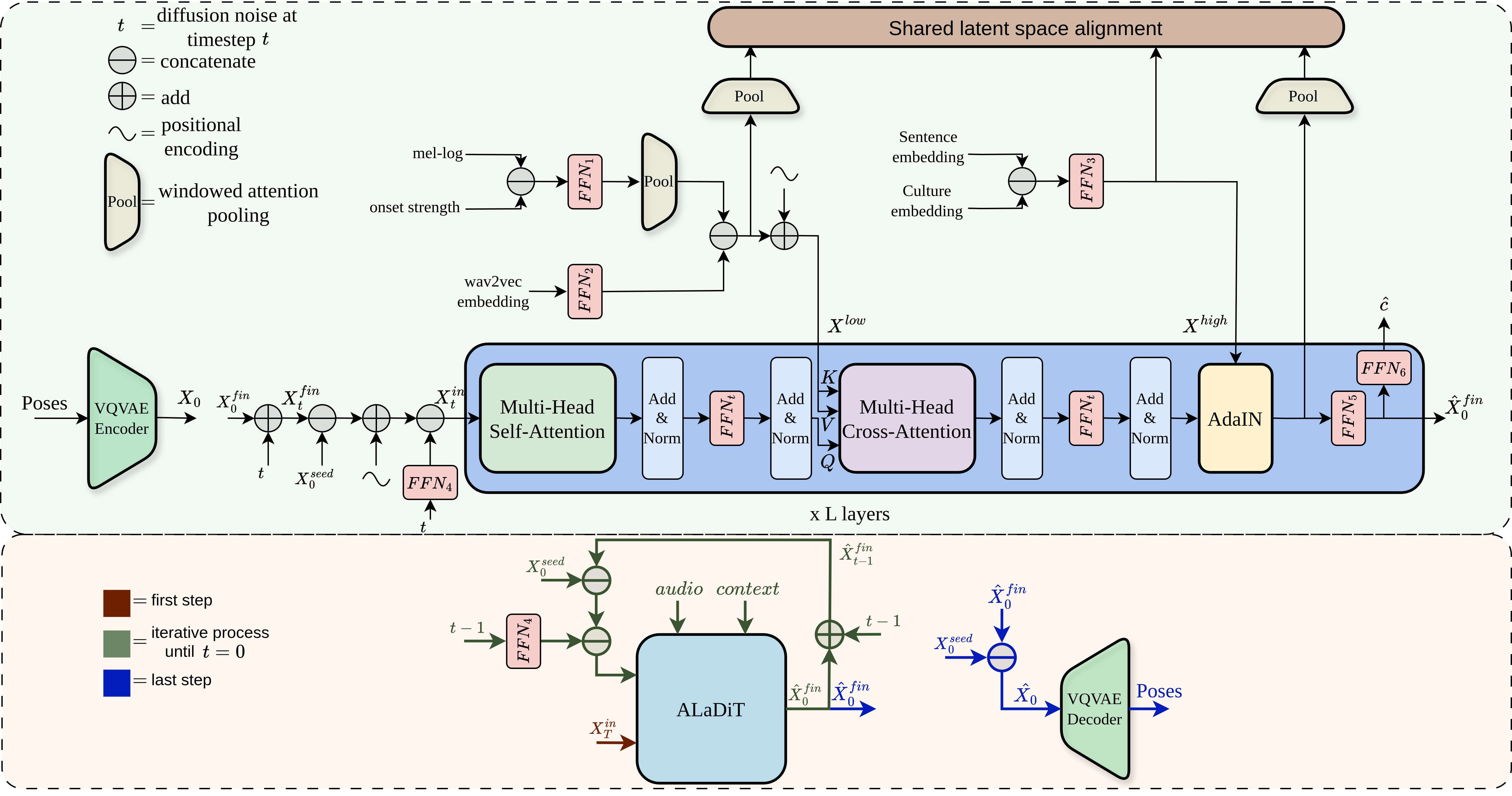}
\caption{\textit{Overview of ALaDiT.} \textbf{Top (Training):} Motion is encoded via a pretrained VQVAE encoder, split into seed $X_0^{seed}$ and target $X_0^{fin}$, which is noised to $X_t^{fin}$. A timestep embedding is concatenated, and motion is processed via self- and cross-attention with audio features ($X^{low}$), while cultural and textual features ($X^{high}$) are injected through AdaIN. All features are aligned in a shared space, while cultural classification $\hat{c}$ further enforces cultural consistency. \textbf{Bottom (Inference)} Given $X_T^{fin}$, $X_0^{seed}$, and context, the model iteratively denoises motion to $\hat{X}_0^{fin}$, then decodes it via the VQVAE decoder.}
\label{fig:SICAGE-model}
\end{figure}

\subsection{Data Collection}

TED4C-L contains YouTube TED Talks from four coarse country-level groups: India (Hindi TED talks), Italy, Japan, and Turkey. These labels serve as practical group annotations and should not be interpreted as implying cultural homogeneity within countries. Our assumption is only that, despite intra-group variability, these groups may contain detectable group-associated communication patterns that can be studied at scale under speaker-disjoint evaluation. We selected groups with distinct spoken languages and a large number of available TED Talks.

To ensure quality, we include only videos in which speakers are clearly visible, standing, speaking their native language (Hindi for India), and not holding objects that could affect gestures. From a total of 106.45h of video data, we extracted 659,454 overlapping five-second samples using 0.5s stride, each with aligned audio, motion, and transcripts. Table~\ref{tab:ted_datasets} compares TED4C-L with the largest available TED-Talk dataset. Although TED4C-L has fewer speakers (764 vs. 1766), it features longer, higher-quality videos and a shorter stride (0.5s vs. 0.67s), resulting in longer duration (106h vs. 97h) and significantly more samples (659k vs. 252k), along with explicit cultural labels, multiple languages, and balanced speaker counts per culture.

Audio is downsampled to 16 kHz, then processed to extract 64 mel-log features (spectral content), onset strength (rhythm), and 1024-dimensional wav2vec features from a model pretrained on 56 languages~\cite{41}. 
For text, we use Language-Agnostic BERT Sentence Embeddings (LaBSE)~\cite{42} to represent contextual meaning, including the first and last words that overlap each sample window for better alignment. 
Since LaBSE mainly aligns sentences by semantic content and may attenuate language-specific idiomatic structure, the audio features provide complementary acoustic, rhythmic, and prosodic cues not captured by text semantics alone.
Motion is represented by 9 upper-body 3D keypoints (neck, head, central hip, shoulders, elbows, wrists) at 15 FPS, extracted with pretrained MMPose~\cite{44} models. Choosing only 9 keypoints reflects a deliberate reliability and scope trade-off. SICAGE targets macro-level upper-body gesticulation in unconstrained TED videos, where hands and fingers are often blurred, occluded, or out of frame, and the lower body is frequently truncated by camera framing. Under these conditions, monocular 3D finger extraction or full-body tracking would introduce substantial detection noise and sharply reduce the amount of usable data. Nonetheless, this compact configuration retains a robust group-level signal, enabling our speaker-disjoint classifier to achieve approximately 45\% balanced accuracy against a 25\% random baseline. 
Following~\cite{51}, we retain only sequences where the main speaker is continuously detected for at least 5 seconds. To reduce camera-dependent variability, we normalize poses to be yaw-invariant (removing average shoulder rotation around the vertical axis) and pitch-invariant (removing hip-to-neck rotation around the horizontal axis), ensuring speakers are front-facing and upright. Each 75-frame motion sequence is represented in 6D continuous format~\cite{45} and encoded by a pretrained VQVAE into 25 tokens (1024-entry VQVAE codebook with 512-dimensional embeddings) via a 1D convolutional encoder with residual blocks and velocity/acceleration losses~\cite{46} to reduce jitter and improve reconstruction (see supplementary for details).

\begin{table}[ht]
\centering
\resizebox{.65\columnwidth}{!}{%
\begin{tabular}{@{}lcc@{}}
\hline
                            & \textbf{TED Talk} & \textbf{TED4C-L} \\
\hline
\textbf{Speakers}                            & 1766              & 764             \\
\textbf{Cultures}                            & /                 & 4               \\
\textbf{Shots of Interest}                   & /                 & 16,326          \\
\textbf{Motion FPS}                          & 15                & 15              \\
\textbf{Duration shots of interest (hours)}  & 97h               & 106.45h          \\
\textbf{Average Video Length (minutes)}      & 13m               & 15.41m          \\
\textbf{Transcript Languages}                & En                & En/Tr/It/Hi/Ja  \\
\textbf{Languages Spoken}                    & /                 & Tr/It/Hi/Ja     \\
\textbf{Sample Duration (frames)}            & 34                & 75              \\
\textbf{Stride (seconds)}                    & 0.67              & 0.5             \\
\textbf{Total Samples}                       & 252,109           & 659,454         \\
\hline
\end{tabular}%
}
\caption{Comparison between TED Talk \cite{51} and TED4C-L Datasets.}
\label{tab:ted_datasets}
\end{table}

\begin{figure}[htbp]
\centering
\includegraphics[width =  .63\columnwidth]{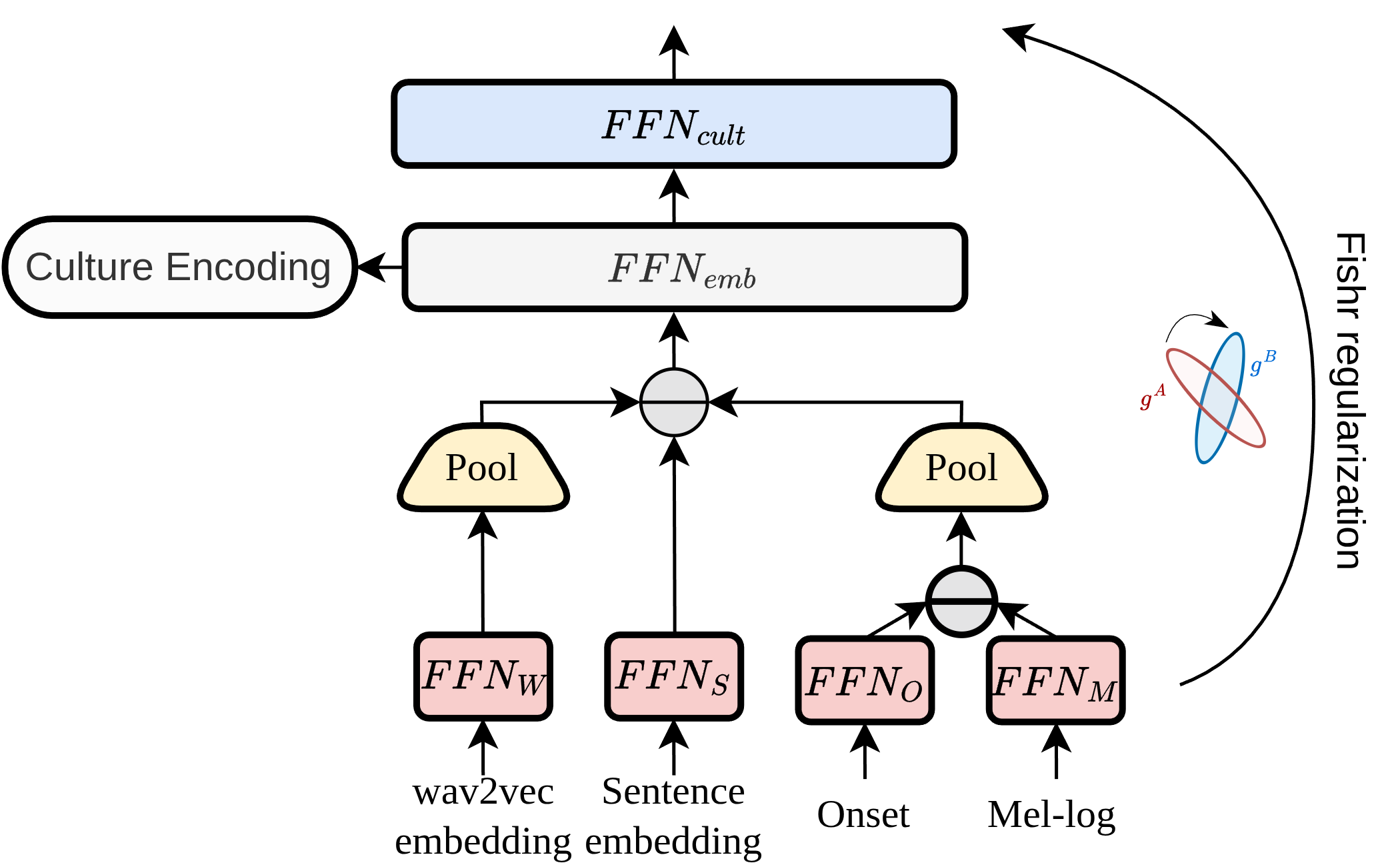}
\caption{Feed-forward network trained with Fishr regularization to learn speaker-independent cultural embeddings.
}
\label{fig:Fishr}
\end{figure}

\subsection{Speaker-Independent cultural representation}
\label{sec:subj-ind}
Cultural traits should generalize across individuals from the same group. To obtain speaker-independent representations, we split the dataset so that training, validation, and test sets contain disjoint speakers within each culture. We then train a feed-forward network (FFN) to classify culture using either Fishr regularization or adversarial learning for domain generalization, treating each speaker as a domain and culture as the prediction target.

For Fishr, due to the large number of domains, we randomly sample $k=64$ speakers per training step. For each selected domain $d \in \mathcal{S}$ (with $\mathcal{S} \subseteq \{1,\dots,D\}$ and $|\mathcal{S}| = 64$), we form a minibatch of $n_d = 16$ samples.

As illustrated in Figure~\ref{fig:Fishr}, audio features (wav2vec, mel-log, and onset) and sentence embeddings are processed by separate $FFN_{(*)}$ branches with attention pooling, concatenated, and projected by $FFN_{emb}$, before being passed to the culture classifier $FFN_{cult}$. Motion is intentionally not used to learn the cultural embedding: target motion is the output to be generated and is therefore unavailable at inference, while the one-second seed is absent at sequence start and too short to estimate culture reliably.

The overall loss for a training step is:
\[
L(\theta) = \frac{1}{N} \sum_{d \in \mathcal{S}} \sum_{i=1}^{n_d} \ell_i^d + \lambda_p\, P(\theta) + \lambda_s\mathcal{L}_{\mathrm{SupCon}}^{(\tau)},
\]
\[ 
\text{where} \; \; \; \; N = \sum_{d \in \mathcal{S}} n_d, \quad
\ell_i^d = -\log p\left(c_i^d \mid x_i^d; \theta\right), 
\]

where $\lambda_p$ is a penalty weight, $c_i^d$ denotes the cultural label of sample $i$ from speaker $d$, and $\theta$ represents the model parameters.

The term $\mathcal{L}_{\mathrm{SupCon}}^{(\tau)}$ represents the supervised contrastive loss \cite{supcon} operating on the learned cultural embeddings to enforce speaker invariance. By considering a multi-domain batch from $1$ to $N$, it is formulated as:
\[
\mathcal{L}_{\mathrm{SupCon}}^{(\tau)} = \frac{1}{N} \sum_{i=1}^{N} \left[ -\frac{1}{|P(i)|} \sum_{p \in P(i)} \log \frac{\exp\left(\mathbf{z}_i^\top \mathbf{z}_p / \tau\right)}{\sum_{a \in A(i)} \exp\left(\mathbf{z}_i^\top \mathbf{z}_a / \tau\right)} \right],
\]

where $\tau$ is the temperature parameter, $A(i) = \{1, \dots, N\} \setminus \{i\}$ represents the set of all sample indices within the current batch excluding the anchor instance itself, and $P(i) = \{p \in A(i) : c_p = c_i\}$ denotes the set of all positive sample indices sharing the same cultural label $c$ as the anchor.

The Fishr penalty $P(\theta)$ quantifies discrepancies in gradient variance across domains:
\[
P(\theta) = \frac{1}{|\mathcal{S}|} \sum_{d \in \mathcal{S}} \| v^d - \bar{v} \|^2
\]
\[
\text{with} \; \; \; \; v^d = \frac{1}{n_d} \sum_{i=1}^{n_d} \left( g_i^d - \bar{g}^d \right)^2, \quad \bar{g}^d = \frac{1}{n_d} \sum_{i=1}^{n_d} g_i^d ,\; \; \; \; \;  g_i^d = \nabla_\theta \ell_i^d
\]
where \(\bar{v} = |\mathcal{S}|^{-1}\sum_{d\in\mathcal{S}} v^d\) denotes the mean variance across domains.
We set $\lambda_p = 0$ for the first 500 updates to stabilize training, and then increase it to $\lambda_p = 1000$ to enforce invariant gradient statistics. The supervised contrastive weight is set to $\lambda_s = 0.2$, and the temperature parameter is $\tau = 0.07$.

For adversarial learning, we use the same architecture as in the Fishr model, with an additional speaker-classification head after a gradient reversal layer (GRL) \cite{49}. The speaker classifier is trained to predict speaker identity, while the gradient reversal layer makes the shared encoder discard speaker-specific information. Let \(\theta_l\), \(\theta_c\), and \(\theta_s\) denote the parameters of the shared encoder, culture classifier, and speaker classifier. The encoder is optimized with:
\[
\mathcal{L}_{\text{tot}}(\theta_l, \theta_{\text{c}}, \theta_{\text{s}}) = \mathcal{L}_{\text{cult}}(\theta_l, \theta_{\text{c}}) -  \lambda_2 \cdot \mathcal{L}_{\text{spk}}(\theta_l, \theta_{\text{s}})  + \lambda_s\mathcal{L}_{\mathrm{SupCon}}^{(\tau)}
\]
The adversarial weight \(\lambda_2\) is gradually increased during training~\cite{49}. The losses $\mathcal{L}_{\text{cult}}$ and $\mathcal{L}_{\text{spk}}$ are the cross-entropy terms for culture and speaker classification.

\subsection{Culture-aware Gesture Generation model}

We propose ALaDiT, a diffusion-based architecture for gesture generation incorporating six modalities: mel-log spectrogram, onset strength, wav2vec embeddings, sentence embeddings, seed motion, and culture embeddings. By using motion embeddings instead of raw poses, our model reduces jitter and focuses on key motion characteristics. ALaDiT can generate a 4-second motion sequence in under 14 ms, enabling real-time gesture synthesis.

As shown in Figure~\ref{fig:SICAGE-model}, each sample comprises 25 motion tokens, split into a 1-second seed ($X_0^{seed} \in \mathbb{R}^{T_p \times d}$, $T_p=5$) and a four-second target ($X_0^{fin} \in \mathbb{R}^{T_r \times d}$, $T_r=20$). Following Motion Diffusion Model (MDM)~\cite{14}, Gaussian noise at timestep $t$ is added to $X_0^{fin}$, yielding $X_t^{fin}$. Each sample also contains five seconds of aligned features: mel-log spectrogram ($X^{mel}$), onset strength ($X^{on}$), wav2vec embeddings ($X^{w2v}$), sentence embedding ($X^{text}$), and culture embedding ($X^{cu}$).

To construct the low-level audio context $X^{low} \in \mathbb{R}^{(T_p + T_r) \times d}$, we concatenate mel-log and onset features and project them to $d/2$ dimensions via $FFN_1$, while wav2vec features are projected to $d/2$ via $FFN_2$. We apply windowed attention pooling to the mel-log and onset features to align their temporal dimension ($T_{mel}$) to that of wav2vec ($T_{w2v}$). The projected features are then concatenated and further downsampled to $(T_p + T_r)$ tokens using another windowed attention pooling step, yielding $X^{low}$, such that each audio token corresponds to a motion token. This pooling operation functions like standard attention-pooling but restricts attention to inputs within a fixed window, efficiently downsampling the temporal dimension while preserving local context.

After applying Positional Encoding ($PE$) to $X^{low}$, we build the high-level context $X^{high}$ by concatenating sentence and culture embeddings, followed by a projection $FFN_3$. The seed $X_0^{seed}$ and noisy motion $X_t^{fin}$ are concatenated, and then positional encoding is added to create $X^{mot}$. The diffusion timestep embedding $e_t$ is computed using sinusoidal encodings followed by a projection $FFN_4$, and then concatenated with $X^{mot}$ to yield:

\[
X^{in} = \mathsf{concat}\Big( \mathbf{e}_t,~X_{mot}\Big), \; \; \; X^{in} \in \mathbb{R}^{(1+T_p+T_r) \times d}
\]

A 10-layer hierarchical Transformer then applies: self-attention to $X^{in}$, cross-attention to $X^{low}$, and finally AdaIN conditioning~\cite{47} with $X^{high}$, producing $X^{out} \in \mathbb{R}^{(1 + T_p + T_r) \times d}$

A residual connection and layer normalization follow each attention layer. After processing, the first six tokens ($e_t$ and $X_0^{seed}$) are discarded. The rest are projected by $FFN_5$ to form the denoised motion $\hat{X}_0^{fin}$. The reconstruction loss is defined as:
\[
\mathcal{L}_{rec} = \mathbb{E}_{\substack{X_0 \sim p(X_0 \mid (X^{low}, X^{high})) \\ t \sim [1, T]}} \left[\mathcal{H} \left(X_0^{fin} - \hat{X}_0^{fin}\right)\right]
\]
where $\mathcal{H}$ is the Huber Loss \cite{54}. To align the generated motion with both low-level audio and high-level textual/cultural contexts, we project $X^{out}$ and $X^{low}$ into a common space of dimension $d$ using windowed attention pooling, resulting in the embeddings $z^{o}$ and $z^{l}$, while $X^{high}$ remains in $\mathbb{R}^d$. We then define the cosine-alignment losses: 
\[
  \mathcal{L}_{low}
  = \mathbb{E}[\,1 - \cos(z_{o},z_{l})]
  ,\;
  \mathcal{L}_{high}
  = \mathbb{E}[\,1 - \cos(z_{o},X^{high})]
\]
To avoid trivial solutions, we further incorporate a contrastive loss \cite{48} to pull together matching pairs and separate non-matching ones.

We apply this loss separately for the low-level (\mbox{$\mathcal{L}_{\text{contrastive}}^{\text{low}}$}) and the high-level (\mbox{$\mathcal{L}_{\text{contrastive}}^{\text{high}}$}) contexts, and define the final contrastive loss  $\mathcal{L}_{cont}$ as their average.

Following \cite{31}, we add a classification head $FFN_6$ on $\hat{X}_0^{fin}$ to predict the culture label $\hat{c}$, and include a cross-entropy loss $\mathcal{L}_{\text{cult}}$.

The final loss is a weighted sum of these loss components:
\[
\mathcal{L}
  = \mathcal{L}_{rec}
  + \lambda_{cu}\,\mathcal{L}_{cult}
  + \lambda_{l}\,\mathcal{L}_{low}
  + \lambda_{h}\,\mathcal{L}_{high}
  + \lambda_{c}\,\mathcal{L}_{cont}
\]

where $\lambda_{cu}$, $\lambda_l$, and $\lambda_h$ are set to $0.1$, and $\lambda_c=0.01$.

\section{Experimental Setup}
\label{sec:exp}

\noindent\textbf{ALaDiT variants.} We train three main ALaDiT variants differing only in cultural conditioning:(i) \textit{ALaDiT NC}, where culture must be inferred implicitly from the audio/text while an auxiliary culture-classification head is used during training;
(ii) \textit{ALaDiT FI}, which is conditioned on speaker-independent cultural embeddings learned with Fishr regularization; and (iii) \textit{ALaDiT ADV}, which is conditioned on embeddings learned via adversarial domain generalization. These variants isolate the effect of explicit culture embeddings and different domain generalization strategies. To isolate possible confounds, we also evaluate three additional ALaDiT ablations: \textit{OneHot}, which conditions on the discrete group label after projecting it with a two-layer MLP and injecting it through the same conditioning pathway; \textit{NoDG}, which uses the same audio/text embedding architecture as FI but removes speaker-domain regularization; and \textit{NoAlign}, which uses Fishr embeddings but removes ALaDiT's multimodal alignment losses.

\noindent\textbf{Baselines comparison.}
We adapt Motion Diffusion Model (MDM)~\cite{14} and DiffuseStyleGesture+ (DSG+)~\cite{50} using the same TED4C-L features as ALaDiT whenever supported by the architecture, while matching optimization settings and diffusion steps ($T=50$). We train NC, FI, and ADV variants for each baseline. For DSG+, we additionally evaluate \textit{DSG+/FI+Align}, which adds an explicit multimodal alignment loss to test whether DSG+ can better exploit Fishr embeddings when alignment is provided. Details on how culture embeddings are injected are provided in the supplementary material.

\noindent\textbf{Objective evaluation.}
We evaluate models on the test split using (i) Fr\'echet Gesture Distance (FGD)~\cite{51}, (ii) Semantic Relevance Gesture Recall (SRGR) \cite{52}, (iii) Beat Alignment Score (BAS) \cite{53}, and (iv) Diversity (mean $\ell_1$ distance between randomly sampled generated motion codebook sequences).
Metrics are averaged over 10 evaluations, each computed on a random subsample of 3000 test instances drawn without replacement; statistical significance is assessed with paired t-tests ($p<0.01$).
To quantify cultural consistency, we apply a speaker-disjoint motion-based cultural classifier to generated gestures and report weighted F1-score on Cultural Expressivity (CE F1) (see supplementary).
SRGR is computed on temporally aligned motion pairs using the original PCK formulation with $\delta=0.05$, BAS uses $\sigma=3$, and FGD embeddings are extracted using the pretrained VQVAE.

\noindent\textbf{Qualitative comparisons.}
To visualize culture-specific differences, we translate the sentence
\textit{``This example helps explain the idea of cultural styles''}
into each dataset language (Google Translate) and synthesize speech with Bark.
Using the same fixed ground-truth seed pose, we generate motions for each culture and compare ALaDiT-FI, ALaDiT-ADV, and ALaDiT-NC.

\noindent\textbf{User study.}
We conduct a user study with $N=20$ participants.
For each culture and condition (Real, NC, FI, ADV), we generate two 30-second clips (32 clips total).
Clips are presented in random order with original audio and English subtitles, and each participant rates all clips.
After each clip, participants evaluate six questions from~\cite{32} (speech coherence, appropriateness, fluency, timing, amount of gesticulation, naturalness) plus an additional question on cultural fit using an 11-point Likert scale (0–10). Before the study, participants viewed two short TED Talk examples per culture to familiarize themselves with the evaluated gesturing styles.

\begin{table}[t]
\centering
\scriptsize
\setlength{\tabcolsep}{2.6pt}
\renewcommand{\arraystretch}{1.08}
\resizebox{\columnwidth}{!}{%
\begin{tabular}{@{}lccccc@{}}
\toprule
\textbf{Model} & \textbf{FGD $\downarrow$} & \textbf{CE F1(\%) $\uparrow$} & \textbf{BAS(\%) $\uparrow$} & \textbf{SRGR(\%) $\uparrow$} & \textbf{Diversity $\uparrow$} \\
\midrule
($\S$) OneHot & $1.63 \pm 0.23$ & $43.73 \pm 1.13$$^{\ast}$ & $22.51 \pm 0.17$ & $67.63 \pm 0.25$ & {\boldmath$111.79 \pm 0.58$}$^{\ddagger,\dagger,\circ}$ \\
($\P$) NoDG & $1.56 \pm 0.22$ & $43.18 \pm 1.20$ & $22.51 \pm 0.23$ & $67.76 \pm 0.23$$^{\ast,\S}$ & $111.60 \pm 0.71$$^{\ddagger,\dagger,\circ}$ \\
($\circ$) NoAlign & $1.36 \pm 0.16$$^{\ddagger,\ast,\S,\P}$ & $43.37 \pm 0.91$ & $22.58 \pm 0.17$ & {\boldmath$68.17 \pm 0.23$}$^{\ddagger,\ast,\S,\P}$ & $111.10 \pm 0.77$$^{\ddagger,\dagger}$ \\
($\ddagger$) NC & $1.60 \pm 0.18$ & $43.41 \pm 1.10$ & $22.51 \pm 0.15$ & $67.72 \pm 0.23$ & $109.50 \pm 0.68$ \\
($\ast$) ADV & $1.53 \pm 0.17$ & $42.71 \pm 0.95$ & $22.45 \pm 0.17$ & $67.57 \pm 0.27$ & $111.75 \pm 0.71$$^{\ddagger,\dagger,\circ}$ \\
($\dagger$) FI & {\boldmath$1.03 \pm 0.15$}$^{\ddagger,\ast,\S,\P,\circ}$ & {\boldmath$44.61 \pm 0.95$}$^{\ddagger,\ast,\S,\P,\circ}$ & {\boldmath$22.63 \pm 0.22$} & $68.09 \pm 0.25$$^{\ddagger,\ast,\S,\P}$ & $110.27 \pm 0.70$$^{\ddagger}$ \\
\bottomrule
\end{tabular}
}
\caption{ALaDiT ablations, reported as mean $\pm$ std over 10 test runs. FI uses Fishr regularization; ADV uses adversarial learning; NC removes explicit cultural conditioning; OneHot uses one hot cultural labels for conditioning; NoDG uses the same audio/text embedding architecture as FI, but without speaker-domain regularization; NoAlign removes ALaDiT alignment losses while using Fishr regularization. Bold marks the best value within the ALaDiT block. Superscripts denote values significantly better than the rows indicated by the corresponding symbols under paired two-sided $t$-tests ($p<0.01$).}
\label{tab:overall_metrics}
\end{table}

\begin{table*}[htbp]
\centering
\scriptsize
\setlength{\tabcolsep}{4pt}
\renewcommand{\arraystretch}{1.02}
\resizebox{\textwidth}{!}{%
\begin{tabular}{@{}lccccc@{}}
\toprule
\textbf{Model} & \textbf{FGD $\downarrow$} & \textbf{CE F1(\%) $\uparrow$} & \textbf{BAS(\%) $\uparrow$} & \textbf{SRGR(\%) $\uparrow$} & \textbf{Diversity $\uparrow$} \\
\midrule
\multicolumn{6}{c}{\textit{DSG+ variants and ablations}} \\
\midrule
($\triangle$) DSG+/NC & $2.76 \pm 0.31$$^{\diamond,\bullet}$ & $41.51 \pm 0.78$$^{\star}$ & $22.48 \pm 0.11$ & $68.17 \pm 0.24$$^{\diamond,\star}$ & $108.85 \pm 0.63$$^{\diamond}$ \\
($\diamond$) DSG+/ADV & $4.81 \pm 0.43$ & {\boldmath$42.21 \pm 0.96$}$^{\triangle,\bullet,\star}$ & $22.58 \pm 0.19$ & $66.78 \pm 0.32$$^{\star}$ & $107.78 \pm 0.71$ \\
($\bullet$) DSG+/FI & $4.89 \pm 0.40$ & $40.67 \pm 1.31$ & $22.48 \pm 0.15$ & {\boldmath$68.46 \pm 0.29$}$^{\triangle,\diamond,\star}$ & $108.85 \pm 1.14$$^{\diamond}$ \\
($\star$) DSG+/FI+Align & {\boldmath$2.52 \pm 0.21$}$^{\diamond,\bullet}$ & $39.80 \pm 0.56$ & {\boldmath$22.67 \pm 0.16$}$^{\triangle}$ & $65.17 \pm 0.24$ & {\boldmath$111.13 \pm 1.07$}$^{\triangle,\diamond,\bullet}$ \\
\midrule
\multicolumn{6}{c}{\textit{MDM variants}} \\
\midrule
($\S$) MDM/NC & $15.58 \pm 1.43$ & $38.57 \pm 0.80$ & $22.52 \pm 0.14$ & $51.62 \pm 0.22$ & $107.62 \pm 1.08$$^{\P}$ \\
($\P$) MDM/ADV & $13.67 \pm 1.17$$^{\S}$ & $38.92 \pm 0.93$ & $22.59 \pm 0.13$ & {\boldmath$52.25 \pm 0.25$}$^{\S,\circ}$ & $105.92 \pm 0.84$ \\
($\circ$) MDM/FI & {\boldmath$7.59 \pm 0.59$}$^{\S,\P}$ & {\boldmath$47.09 \pm 0.79$}$^{\S,\P}$ & {\boldmath$22.59 \pm 0.17$} & $51.86 \pm 0.24$ & {\boldmath$109.37 \pm 0.74$}$^{\S,\P}$ \\
\midrule
\multicolumn{6}{c}{\textit{ALaDiT variants}} \\
\midrule
($\ddagger$) ALaDiT/NC & $1.60 \pm 0.18$ & $43.41 \pm 1.10$ & $22.51 \pm 0.15$ & $67.72 \pm 0.23$ & $109.50 \pm 0.68$ \\
($\ast$) ALaDiT/ADV & $1.53 \pm 0.17$ & $42.71 \pm 0.95$ & $22.45 \pm 0.17$ & $67.57 \pm 0.27$ & {\boldmath$111.75 \pm 0.71$}$^{\ddagger,\dagger}$ \\
($\dagger$) \textbf{ALaDiT/FI (ours)} & {\boldmath$1.03 \pm 0.15$}$^{\ddagger,\ast}$ & {\boldmath$44.61 \pm 0.95$}$^{\ddagger,\ast}$ & {\boldmath$22.63 \pm 0.22$} & {\boldmath$68.09 \pm 0.25$}$^{\ddagger,\ast}$ & $110.27 \pm 0.70$$^{\ddagger}$ \\
\bottomrule
\end{tabular}%
}
\caption{Comparison across DSG+, MDM, and ALaDiT variants (FI, ADV, NC), reported as mean $\pm$ std over 10 matched test runs. 
Bold marks the best value within each architecture family, not necessarily the best global value. Superscripts denote significantly better results than the marked row within the same family under paired two-sided $t$-tests ($p<0.01$).}
\label{tab:model_comparison}
\end{table*}

\section{Results}
\label{sec:res}

\subsection{Quantitative analysis}

We evaluate SICAGE through both ablation studies and comparisons with strong diffusion-based gesture generation baselines.

\noindent\textbf{Effect of speaker-independent cultural embeddings.}
Table~\ref{tab:overall_metrics} reports an ablation study of the proposed cultural representation within the same ALaDiT generator. Introducing explicit cultural representations improves most metrics. Fishr conditioning (FI) achieves the best motion realism (lowest FGD) and the highest cultural classification accuracy (CE F1), indicating that speaker-invariant embeddings learned with Fishr capture culturally relevant motion patterns more effectively. Adversarial learning (ADV) remains closer to NC on most metrics, except for Diversity, suggesting that the adversarial objective provides weaker perceptual and objective gains in this setting. 
The OneHot ablation shows that a discrete cultural label alone is not sufficient: although it can increase Diversity, it does not match FI in FGD or CE F1. NoDG further shows that using the same audio/text embedding architecture without speaker-domain regularization is also insufficient. Finally, NoAlign confirms that ALaDiT's alignment losses help exploit the learned embeddings, but do not by themselves explain the FI gains: removing alignment worsens FGD and CE F1 despite a small increase in SRGR. Overall, the ablations show that the strongest results come from combining Fishr-based speaker-independent embeddings with an architecture able to align multimodal conditioning.

\noindent\textbf{Comparison with existing models.}
Table~\ref{tab:model_comparison} compares ALaDiT with DiffuseStyleGesture+ (DSG+) and Motion Diffusion Model (MDM), each trained with the same features and diffusion settings. Overall, ALaDiT provides the strongest trade-off across the evaluated metrics. In particular, ALaDiT achieves substantially lower FGD, indicating closer similarity to real motion distributions, while maintaining competitive or superior CE F1, BAS, Diversity, and SRGR scores. 
Introducing cultural embeddings generally improves performance for the baseline models. For MDM, both FI and ADV improve several metrics, particularly FGD and CE F1, with FI consistently outperforming ADV, following the same trend observed for ALaDiT.  In contrast, improvements for DSG+ are more limited, suggesting that cultural embeddings are most effective when combined with architectures that explicitly model multimodal alignment. This is further supported by the DSG+/FI+Align ablation, where adding explicit alignment substantially improves FGD, Diversity, and BAS over DSG+/FI, although performance still remains below ALaDiT/FI in terms of motion realism and cultural consistency.
Taken together, these results show that (i) speaker-independent cultural embeddings significantly improve gesture generation when effectively integrated within the generator architecture, (ii) Fishr provides the strongest cultural representation among the evaluated domain-generalization strategies, and (iii) the proposed ALaDiT architecture achieves the best overall performance when combined with these embeddings.

\begin{figure}[htbp] 
\centering 
\includegraphics[width=\columnwidth]{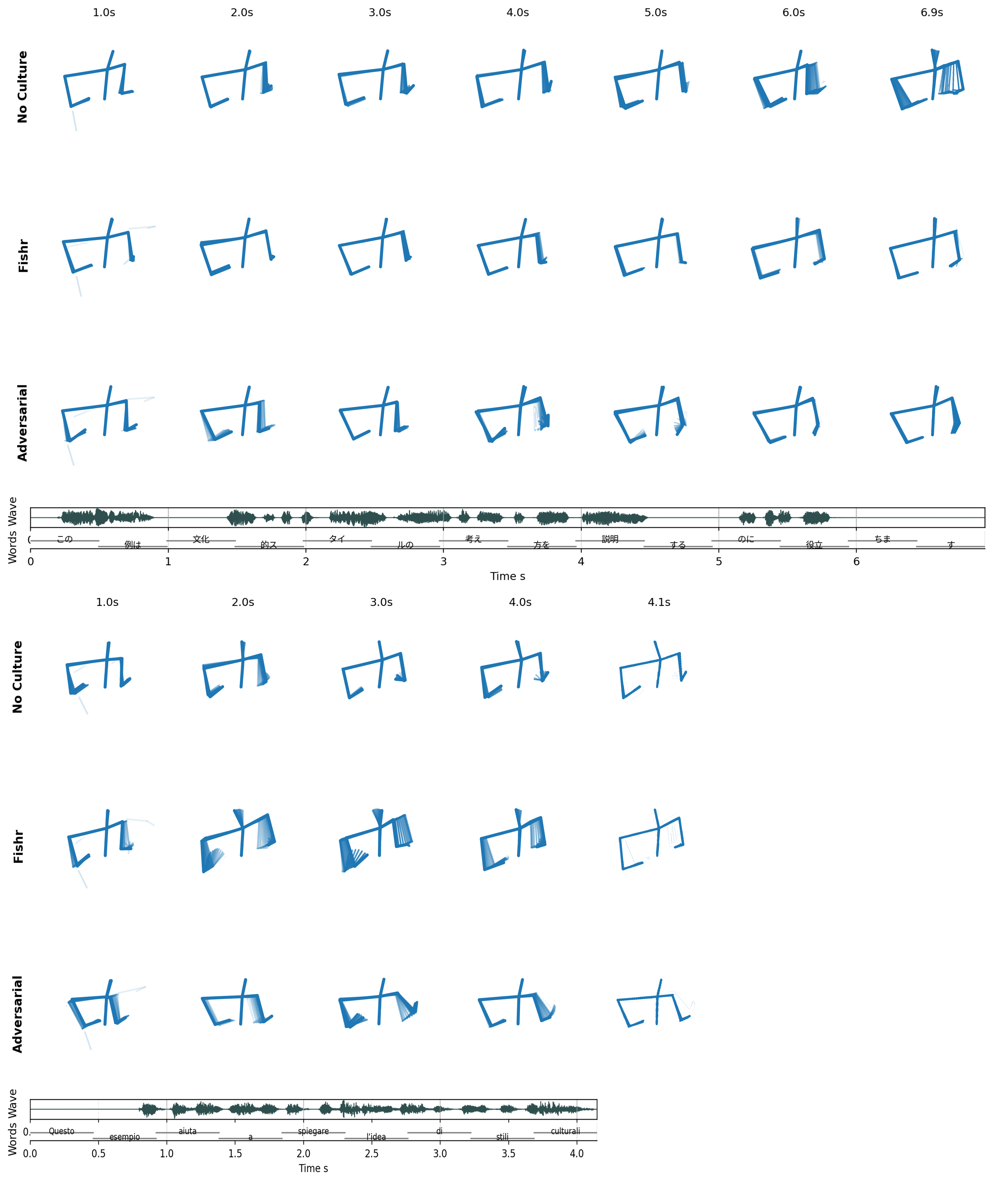}
\caption{Motion generated by the No Culture, Fishr and Adversarial models for Japanese (top rows) and Italian (bottom rows) cultures given the sentence ``This example helps explain the idea of cultural styles''. Each image represents one second of motion, except the last one, which represents the last frames.}
\label{fig:qualitative}
\end{figure}

\subsection{Qualitative analysis}

We visually compare ALaDiT-FI and ALaDiT-ADV with the no-culture variant (NC) under matched seed pose and semantic content (``This example helps explain the idea of cultural styles''), translated into each language and synthesized with Bark. Figure~\ref{fig:qualitative} shows Japanese and Italian examples, where each column summarizes one second of motion at 15 fps, while the final column shows the remaining frames of the sequence; darker regions indicate joint locations that are occupied more frequently within that second.
For Japanese, all models produce relatively compact upper-body motion throughout the utterance. However, NC exhibits higher spatial dispersion at the wrists and elbows, particularly toward the end of the clip when speech activity decreases. This residual motion during silence suggests lower motion realism, consistent with the quantitative results. In contrast, FI and ADV show more stable motion trajectories and terminate gestures more cleanly as the speech ends.
For Italian, FI, and ADV generate broader spatial gestures and larger arm excursions than NC, with movement sustained across multiple seconds. In contrast, NC remains more constrained and shows a limited range of motion. Among the culture-conditioned variants, FI distributes motion more globally across the upper body, whereas ADV concentrates motion more strongly on the arms (notably the left arm in this example). Overall, these qualitative trends support the benefit of explicit culture embeddings in producing motion that is both better aligned to speech dynamics and more consistent with culture-specific patterns captured by our motion-based classifier. Additional qualitative examples and rendered videos for all cultures, including comparisons to ground-truth motion, are provided in the supplementary material.

\subsection{User study}
\begin{figure}[htbp] 
\centering 
\includegraphics[width=1\columnwidth]{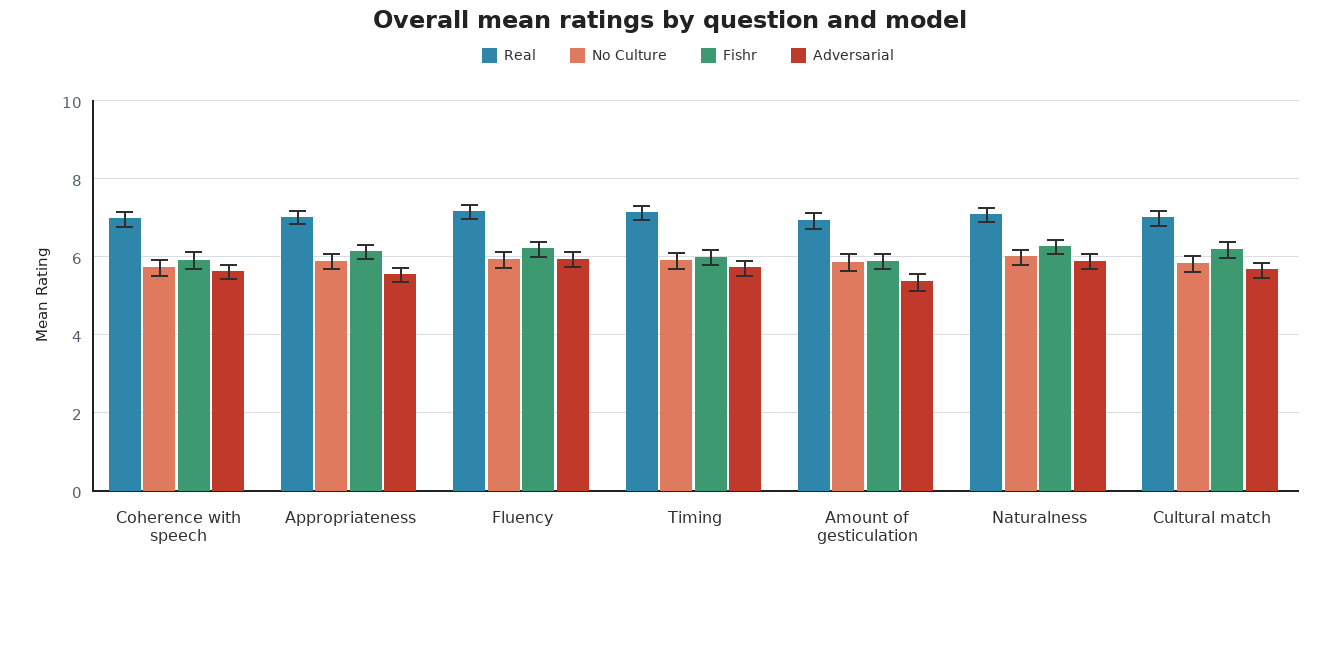}
\caption{User-study scores across cultures and conditions. Bars show mean ratings across participants (N = 20); error bars show standard deviation.}
\label{fig:user_study}
\end{figure}

The user study results shown in Figure~\ref{fig:user_study}, conducted with 20 participants from different cultural backgrounds, further support the findings of the objective and qualitative analyses.
Overall differences between generated models are relatively small, which is expected given the subjective nature of the task and the limited number of raters. Nevertheless, FI obtains the highest average score among generated models and is significantly preferred over ADV overall ($6.06$ vs. $5.65$, $p=0.033$). FI is also significantly preferred over NC on Cultural Match ($6.16$ vs. $5.81$, $p=0.038$), suggesting that Fishr-based cultural embeddings produce perceptible improvements in culture-associated gesture style. 
NC and ADV receive comparable ratings, indicating that the adversarial domain-generalization objective yields less consistent perceptual gains. Real motion remains the highest-rated condition, confirming that generated gestures are still perceptually distinguishable from ground-truth motion.
These findings are consistent with the objective results, where FI provides the strongest balance between motion realism and cultural consistency. We also observe variation across cultures; for Turkish samples, NC is slightly preferred over FI and ADV, but this difference is not statistically significant. A more detailed per-culture analysis is provided in the supplementary material.

\section{Conclusion}
\label{sec:conc}
We introduced SICAGE, a modular framework for culture-aware co-speech gesture generation that learns speaker-independent cultural embeddings by treating speakers as domains. We also introduced TED4C-L, a large-scale multicultural dataset with speaker-disjoint splits for evaluating cultural generalization. Experiments show that explicitly modeling culture improves gesture synthesis. Fishr-based embeddings provide the most consistent gains in motion realism, cultural consistency, and speech–gesture alignment. Combined with these embeddings, our ALaDiT diffusion model achieves state-of-the-art performance compared to recent diffusion-based baselines. Quantitative results, qualitative analysis, and a user study indicate that differences between models are perceptible to human observers, with the Fishr-based model receiving the highest ratings in most evaluation criteria. 
Future work should consider finer-grained cultural annotations, reliable hand/finger tracking in in-the-wild videos, and representation learning methods that better disentangle culture-associated regularities from language, prosody, semantics, and coarse country-level grouping effects. User studies involving participants more familiar with the evaluated cultures could also provide more reliable perceptual assessments.


%
%
\bibliographystyle{splncs04}
\bibliography{main}

\clearpage

\appendix

\setcounter{section}{0}
\setcounter{subsection}{0}
\setcounter{figure}{0}
\setcounter{table}{0}
\setcounter{equation}{0}

\renewcommand{\thesection}{\Alph{section}}
\renewcommand{\thesubsection}{\Alph{section}.\arabic{subsection}}
\renewcommand{\thefigure}{S\arabic{figure}}
\renewcommand{\thetable}{S\arabic{table}}
\renewcommand{\theequation}{S\arabic{equation}}

\renewcommand{\theHsection}{supp.\Alph{section}}
\renewcommand{\theHsubsection}{supp.\Alph{section}.\arabic{subsection}}
\renewcommand{\theHfigure}{supp.\arabic{figure}}
\renewcommand{\theHtable}{supp.\arabic{table}}
\renewcommand{\theHequation}{supp.\arabic{equation}}

\begin{center}
{\Large\bfseries Supplementary Material}\\[1mm]
\end{center}

\vspace{2mm}

\section{VQVAE}
\subsection{Architecture}

Recent work~\cite{58,46} has shown that discrete motion representations via Vector-Quantized Variational Autoencoders (VQVAEs) can improve co-speech gesture generation. By discretizing motion into codebook units, VQVAEs mitigate high-frequency jitter, stabilize training, and yield higher-fidelity reconstructions than conventional continuous (e.g., 6D rotations) representations. In this work, we develop a VQVAE tailored to the TED4C-L dataset, which contains 5-second gesture sequences sampled at 15~fps across diverse speakers and cultures. Our goal is to produce robust, compressed motion representations that facilitate gesture generation, improve motion realism, and generalize across all speakers and cultures in TED4C-L.

\begin{figure} [htbp]
\centering
\includegraphics[width = .7\columnwidth]{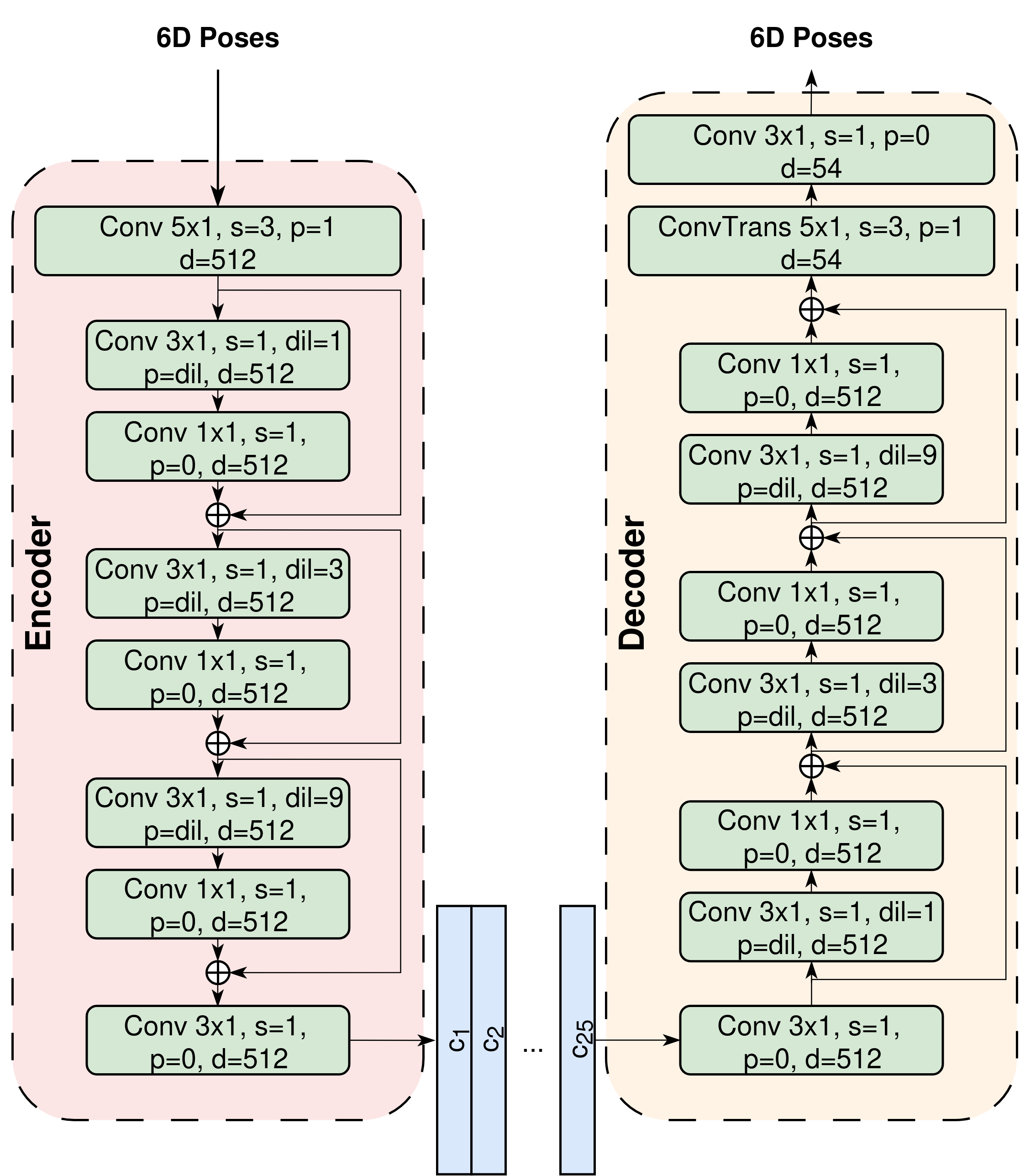}
\caption{VQVAE architecture for reconstructing motion in TED4C-L. Encoder (red), codebooks (blue), and decoder (orange) are highlighted.}
\label{fig:vqvae-scheme-ted}
\end{figure}

Our architecture (Figure~\ref{fig:vqvae-scheme-ted}) is designed to (1) capture both local and long-range temporal dependencies, (2) compress motion efficiently, and (3) reconstruct high-quality gestures. The processing pipeline is:

\begin{enumerate}
\item Input: each gesture sequence consists of 5~seconds of motion ($75$ frames at $15$~fps), with $9$ joints per frame and $6$D rotation per joint, forming a $[75,54]$ matrix.
\item Temporal Compression: a 1D convolutional layer (kernel size $k=5$, stride $s=3$, padding $p=1$, number of filters $d=512$) projects the input to a $[25,512]$ representation, providing the initial compression and contextualization over time.
\item Dilated Convolutional Blocks: three sequential blocks, each with a 1D convolution ($k=3$, $dil \in {1,3,9}$, $p=dil$, $s=1$, $d=512$) followed by another 1D convolution ($k=1$, $p=0$, $s=1$, $d=512$). Dilation rates $dil$ expand the temporal receptive field, enabling the model to capture both short- and long-term dependencies crucial for natural gesture synthesis. Residual connections and padding preserve sequence length and ease optimization.
\item Encoder Output: a final 1D convolution ($k=3$, $p=1$, $d=512$) produces the encoder output.
\item Quantization: each of the $25$ latent vectors is quantized by selecting the nearest entry from a codebook of $1024$ embeddings of dimension $512$, following the VQVAE paradigm. This discrete bottleneck is key to suppressing spurious frame-to-frame noise and focusing the generative model on salient dynamics.
\item Decoder: the decoder mirrors the encoder, using transposed 1D convolutions to progressively expand the temporal dimension. The final output is projected back to $[75,54]$ via a 1D convolution ($k=3$, $s=1$, $p=1$, $d=54$) to match the input dimension and improve local detail in the reconstruction.
\end{enumerate}

The architecture achieves a temporal downsampling factor of 3, balancing compactness and motion fidelity.

To further improve temporal smoothness and physical plausibility, we extend the standard VQVAE loss with additional velocity and acceleration terms, following best practices from prior work \cite{46}. The final loss consists of:  (i) reconstruction loss  $\mathcal{L}_{rec} =
    \big\|x - \hat{x}\big\|_1$, (ii) commitment loss $ \mathcal{L}_{com} =\big\|z_e(x) - \operatorname{sg}(z_q(x))\big\|_2^2$, (iii) velocity loss $ \mathcal{L}_{vel} = \big\|\dot x - \dot{\hat x}\big\|_1$, (iv) acceleration loss $\mathcal{L}_{acc} = \big\|\ddot x - \ddot{\hat x}\big\|_1$, and (v) a temporal regularization term $
\mathcal{L}_{reg}
=
\frac{1}{N(T-2)}
\sum_{n=1}^{N}
\sum_{t=2}^{T-1}
\left\|
\hat{x}^{(n)}_{t+1}
+
\hat{x}^{(n)}_{t-1}
-
2\hat{x}^{(n)}_{t}
\right\|_2^2
$, where $N$ denotes the batch size, $T$ the number of frames in each motion sequence, $x$ and $\hat{x}$ denote the ground-truth and reconstructed motions, $z_e(x)$ is the encoder output, $z_q(x)$ is the quantized representation of the encoder output, $\operatorname{sg}$ is the stop-gradient operation \cite{40}, $\dot{x}$ and $\ddot{x}$ are angular velocities and accelerations. $\mathcal{L}_{reg}$ penalizes the squared second-order finite difference of the reconstructed motion sequence and therefore reduces high-frequency jitter. In the exponential moving average (EMA) variant of VQ-VAE, the codebook embeddings are not optimized through an explicit codebook loss. Instead, each codebook vector is updated using an exponential moving average of the encoder outputs assigned to that entry during training. Note that 
$\mathcal{L}_{acc}$ and $\mathcal{L}_{reg}$ serve complementary purposes. The acceleration loss matches the second-order dynamics of the reconstruction to those of the ground-truth motion, whereas the regularization term directly penalizes excessive second-order variation in the reconstructed sequence itself, encouraging smoother outputs even when the target motion contains local noise.
    
The total loss is:
\[
\mathcal{L} = \mathcal{L}_{rec} + \lambda_{\beta}\mathcal{L}_{com}
+ \lambda_{vel}\mathcal{L}_{vel}
+ \lambda_{acc}\mathcal{L}_{acc}
+ \lambda_{reg}\mathcal{L}_{reg}
\]

Hyperparameters are set as $\lambda_\beta=0.2$, $\lambda_{\text{vel}}=0.1$, $\lambda_{\text{acc}}=0.1$, $\lambda_{reg}=0.1$.

\subsection{Training and evaluation}
We train the VQ-VAE for $300$ epochs on speaker-dependent splits. We use:
\begin{itemize}
    \item Optimizer: Adam ($\beta_1 = 0.9$, $\beta_2 = 0.999$).
    \item Batch size: $512$
    \item Learning rate: $1 \times 10^{-4}$, reduced by $10\times$ every $100$ epochs.
    \item EMA: Exponential Moving Average with $\beta=0.99$ for codebook updates.
    \item Hardware: Single RTX 3090 GPU with 64GB of RAM.
\end{itemize}

Model selection is based on the best mean Euclidean reconstruction error per epoch. Qualitative inspection confirms that the VQVAE achieves smooth, high-fidelity reconstructions that closely match real motion.

\begin{figure}[htbp] 
\centering
\includegraphics[width = .7\columnwidth]{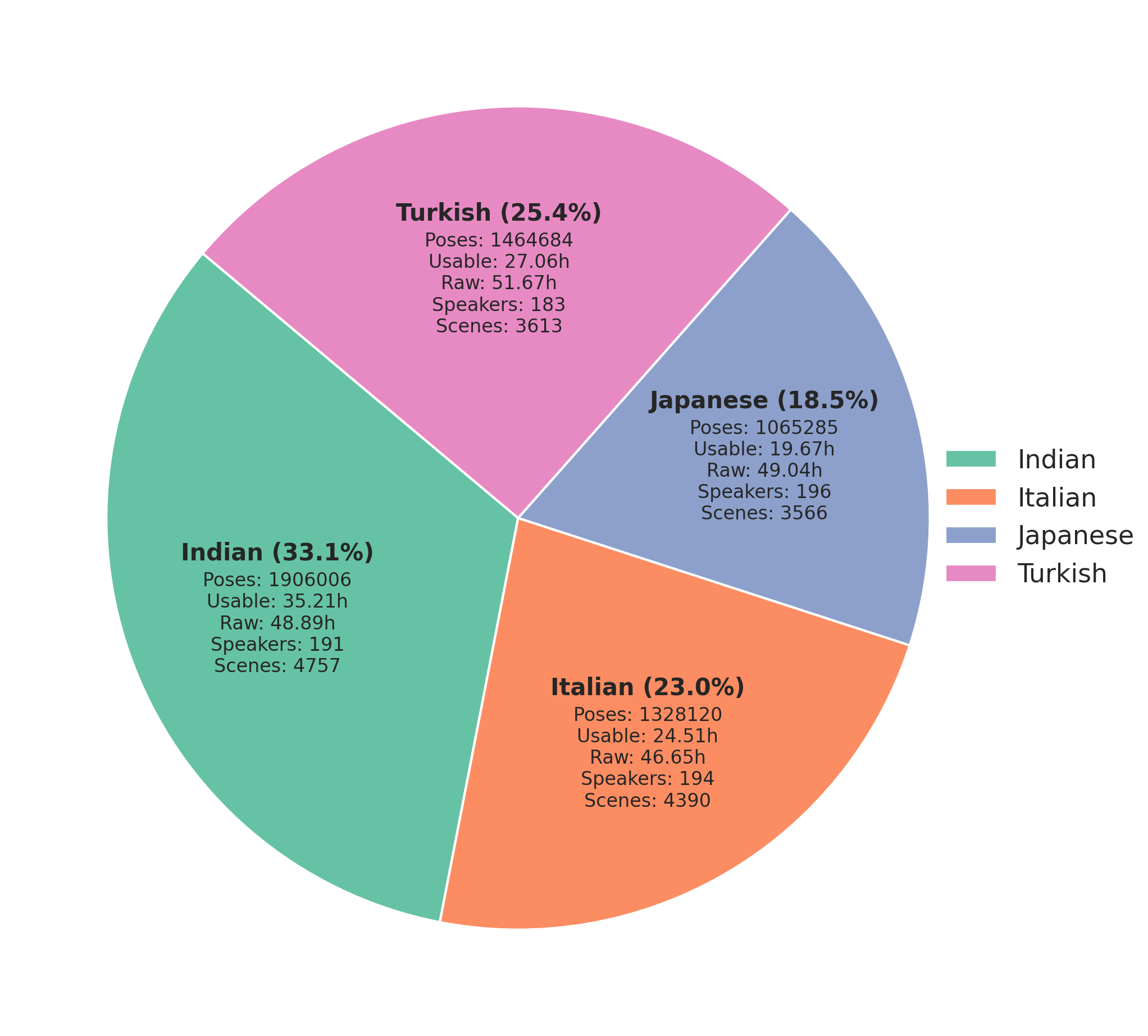}
\caption{TED4C-L data distribution by culture. ``Tot duration” is cumulative video duration per culture; ``Usable" is the subset with reliable pose extraction. ``N.poses," ``N. speakers," and “N. scenes" denote total poses, speakers, and scenes per culture.}
\label{fig:3D-pie}
\end{figure}

\section{Dataset distribution}

This section provides additional details about the TED4C-L dataset. Figure~\ref{fig:3D-pie} summarizes the distribution of data across cultures in terms of cumulative video duration, duration of usable data, number of extracted poses, number of speakers, and number of scenes.

Overall, the dataset is reasonably balanced in terms of speakers, but it is not perfectly balanced in terms of usable motion data. In particular, the Japanese subset contains noticeably fewer extracted poses than the other cultures, despite having a relatively high number of speakers. This suggests that speaker count alone does not determine the final amount of usable pose data. Instead, the difference appears to be related to the visibility of the main speaker and the structure of the scene: in the Japanese subset, the main speaker is less consistently visible, which reduces both the number and duration of valid scenes, and the number of reliable extracted keypoints.

\begin{figure}[htbp] 
\centering
\includegraphics[width = .7\columnwidth]{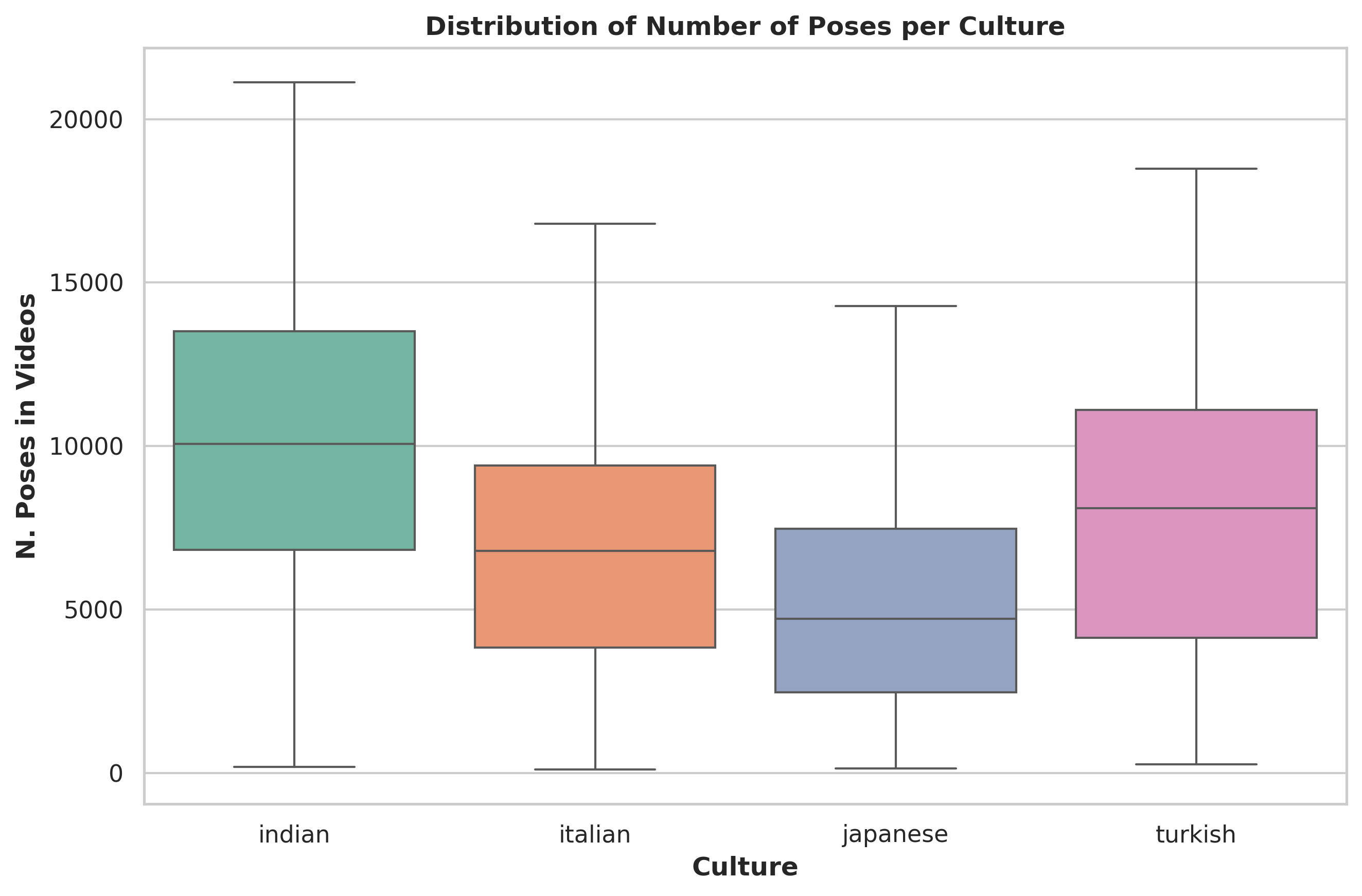}
\caption{Box plot comparing the average number of poses that could be extracted from videos across cultures. The vertical lines indicate standard deviations, the horizontal lines denote median values, and the boxes represent the interquartile range (Q1 to Q3).}
\label{fig:boxplot_poses}
\end{figure}

\begin{figure}[htbp] 
\centering
\includegraphics[width = .7\columnwidth]{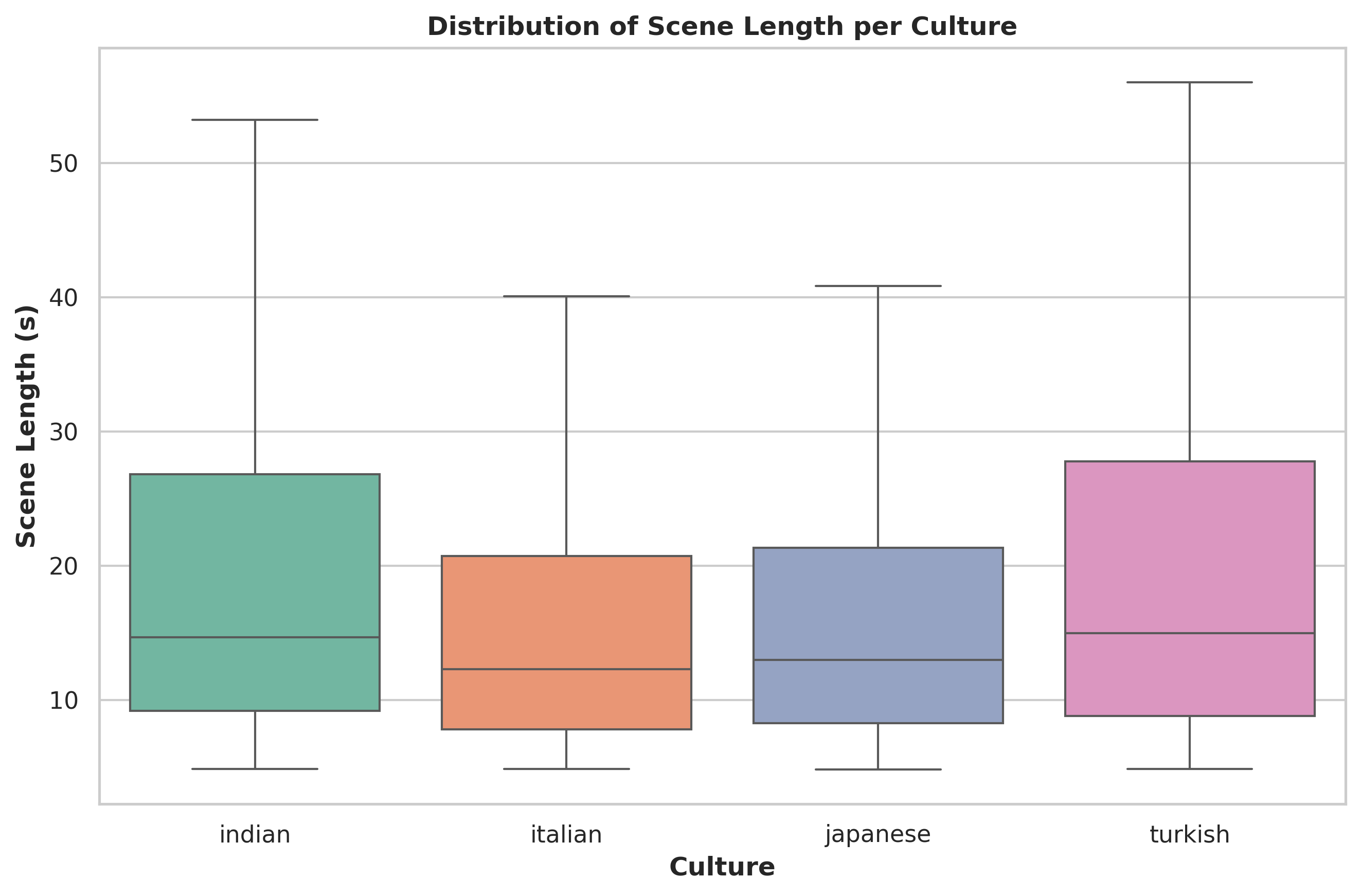}
\caption{Box plot comparing the average scene length (in seconds) of scenes where the main speaker is clearly visible in the videos across cultures. The vertical lines indicate standard deviations, the horizontal lines denote median values, and the boxes represent the interquartile range (Q1 to Q3).}
\label{fig:boxplot_scenes}
\end{figure}

Figure~\ref{fig:boxplot_poses} analyzes this issue by showing the distribution of the average number of poses extracted per video across cultures. Figure~\ref{fig:boxplot_scenes} complements this analysis by showing the distribution of the average length of the scene, measured only in scenes where the main speaker is clearly visible. Together, these plots clarify that the effective amount of usable motion data varies not only with the number of videos or speakers, but also with how often and how long the speaker remains visible on screen.

This analysis is useful for interpreting dataset-level imbalances and can inform future dataset curation.

\section{Generative Models training and evaluation}

We train the proposed ALaDiT model together with the DSG+ \cite{50} and MDM \cite{14} baselines on speaker-independent splits, ensuring no speaker overlap between training, validation, and test sets. For all reported model families, checkpoints are saved every $50{,}000$ steps, and model selection is performed over checkpoints up to the $500{,}000$-step checkpoint.

We use:

\begin{itemize}
    \item Optimizer: AdamW
    \item Batch size: 64
    \item Diffusion steps: 50
    \item Attention heads: 8 
    \item Layers: 10
    \item Latent dimension: 512
    \item Learning rate: $5 \times 10^{-5}$
    \item Hardware: Single RTX 3090 GPU with 64GB of RAM
\end{itemize}

In ALaDiT, the optimizer uses $\beta_1=0.9$ and $\beta_2=0.999$, model parameters are tracked with EMA ($\beta=0.999$), and the learning rate is decayed by a factor of $0.5$ every $100{,}000$ steps. DSG+ and MDM baselines follow their original training code and do not use EMA.

The feed-forward network (FFN) modules in ALaDiT are structured as follows:

\begin{itemize}
\item Each $FFN_{(*)}$, except $FFN_t$ and $FFN_6$, is implemented as a single linear layer followed by GELU activation, Layer Normalization, and dropout (rate $0.1$).
\item $FFN_t$ consists of two linear projections: $512 \rightarrow 2048$, followed by GELU and dropout ($0.1$), and $2048 \rightarrow 512$.
\item $FFN_6$ includes an attention pooling layer (attention weights computed by a single linear projection and softmax), followed by Batch Normalization, dropout, and a final classification layer for culture prediction.
\end{itemize}

For DSG+, culture is treated as a style feature, and all features (wav2vec, onset strength, mel-log spectrogram, sentence embeddings) are concatenated after being mapped to the same temporal length as motion. Attention pooling is used for audio features, and sentence embeddings are repeated along the temporal dimension. For MDM, text and culture embeddings are concatenated, projected to the model’s latent dimension (512), and added to the timestep encoding as in the original implementation. All other hyperparameters follow the original works.

All saved configurations use seed $10$. In ALaDiT training and in the evaluation, Python, NumPy, and PyTorch random seeds are explicitly fixed.

For validation, checkpoints saved every $50{,}000$ steps are evaluated offline on the validation split. For each checkpoint, $10{,}000$ validation samples are synthesized and the overall FGD score is computed; the checkpoint with the lowest validation FGD is selected. The selected checkpoint is then evaluated on the speaker-independent test split over 10 repeated runs. In each run, a seeded subset of $3{,}000$ test instances is sampled, generation is performed on that subset, and all metrics are computed. Final scores are reported as mean $\pm$ standard deviation across the 10 runs. Diversity is computed as the average pairwise Euclidean distance between 300 randomly sampled pairs of flattened continuous motion embeddings. SRGR is computed on temporally aligned motion pairs using the original PCK formulation with $\delta=0.05$, following~\cite{52}. BAS is computed as in~\cite{53} with $\sigma=3$, aligning motion-beat times (local minima in joint-velocity) with audio-beat times (onset strength). FGD is computed in the continuous VQ-VAE latent space by fitting multivariate Gaussians to real and generated motion encodings. The total parameter count for ALaDiT is approximately $50$ million.

\section{Culture Classifiers training and evaluation}

We train the culture classifiers on speaker independent splits, ensuring no speaker overlap between training, validation, and test sets. The multimodal classifiers used to produce the culture embeddings for SICAGE operate only on text and audio features, since motion is not available at inference time. In the reported setting, each sample is represented by sentence embeddings, mel-log spectrogram features, onset-strength features, and wav2vec features.

The final multimodal Fishr and adversarial classifiers share the same encoder backbone. The sentence branch $FFN_S$ maps the $768$-dimensional sentence embedding through two linear projections, $768 \rightarrow 512$ and $512 \rightarrow 512$, each followed by Layer Normalization, GELU activation, and dropout (rate $0.1$). The mel branch $FFN_M$ maps each $64$-dimensional mel frame through $64 \rightarrow 512 \rightarrow 256$ with the same pattern, while the onset branch $FFN_O$ maps each scalar onset value through $1 \rightarrow 512 \rightarrow 256$ with the same structure. The mel and onset features are then concatenated frame-wise to obtain a $512$-dimensional audio sequence over $156$ time steps and summarized with attention pooling, where attention weights are computed by a single linear projection followed by a softmax over time. The wav2vec branch $FFN_W$ maps each $1024$-dimensional wav2vec frame through $1024 \rightarrow 512 \rightarrow 512$, again with Layer Normalization, GELU, and dropout ($0.1$), and the resulting sequence is summarized by attention pooling.

The three modality embeddings, namely the sentence embedding, the pooled mel-onset embedding, and the pooled wav2vec embedding, are concatenated into a $1536$-dimensional vector. This vector is processed by a shared fusion block $FFN_{emb}$ implemented as $1536 \rightarrow 512$, followed by Layer Normalization, GELU, and dropout ($0.1$). A second embedding block maps $512 \rightarrow 512$ with the same LayerNorm-GELU-dropout structure, yielding the final $512$-dimensional culture embedding. This embedding is normalized before classification and is the representation used to condition the generative models.

For Fishr, the normalized embedding is fed to a single linear classification head $FFN_{cult}$ that predicts the cultural class. The culture head $FFN_{cult}$ is implemented as $512 \rightarrow 512 \rightarrow 4$, with Layer Normalization, GELU, and dropout ($0.1$) between the two linear layers. Fishr regularization is applied across speaker domains by matching the gradient-variance statistics of the classifier across speakers; these per-domain statistics are tracked with an exponential moving average of $0.95$. Thus, in the reported multimodal setting, the Fishr model differs from the adversarial model mainly in the training objective and in the final prediction head.

For the adversarial classifier, the same normalized $512$-dimensional embedding is connected to two heads. The speaker-classification head $FFN_{spk}$ has the same structure, with the final layer mapping to the number of training speakers. A gradient reversal layer is applied before $FFN_{spk}$ so that the shared embedding remains discriminative for culture while suppressing speaker-specific information.

To compute Cultural Expressivity (CE) on generated motion, we additionally train a motion-only Fishr classifier. This model takes as input the last $20$ VQ-VAE codebooks of each sample, excluding the first $5$ codebooks corresponding to ALaDiT's seed motion. The resulting $20 \times 512$ sequence is processed by a Transformer encoder with model dimension $512$, $8$ attention heads, $2$ encoder layers, dropout $0.1$, and feed-forward dimension $512$. The transformer output is then projected through a final $512 \rightarrow 512$ linear layer with GELU and dropout ($0.1$), summarized by attention pooling, and classified by a linear culture head. We also trained an adversarial motion-only classifier, but it achieved lower performance on the speaker-independent test set and less stable training; therefore, the Fishr motion-only classifier is the one used to evaluate CE on generated motion.

We train all classifiers for $50$ epochs on speaker-independent splits. We use:
\begin{itemize}
\item Optimizer: AdamW ($\beta_1 = 0.9$, $\beta_2 = 0.999$; weight decay $10^{-4}$)
\item Batch size: $256$ for adversarial training; effective batch size $1024$ for Fishr ($64$ speaker domains $\times 16$ samples per domain)
\item Learning rate: $1\times10^{-4}$
\item Learning rate schedule: cosine annealing for adversarial training with $\eta_{\text{min}} = 1\times10^{-6}$ \cite{59}; no learning-rate scheduler is used for Fishr
\item Fishr regularization: penalty weight $1000$, activated after $500$ warmup updates and linearly ramped over the next $100$ updates
\item Hardware: Single RTX 3090 GPU with 64GB of RAM
\end{itemize}

Validation is performed at every epoch. For both adversarial and Fishr training, we report weighted F1, balanced accuracy, accuracy, ROC-AUC, and cross-entropy on the validation split, and select the checkpoint with the lowest validation cross-entropy. On unseen speakers, both multimodal classifiers achieve approximately $98\%$ weighted F1 for culture recognition. The motion-only Fishr classifier achieves approximately $45\%$ weighted F1, while the adversarial motion-only classifier reaches approximately $40.5\%$. For this reason, and because of its more stable training behavior, the Fishr motion-only classifier is used to compute CE F1 on generated motion.

\section{Per-culture results}
\label{sec:per_culture_objective_subjective_discussion}

\begin{table}
\centering
\scriptsize
\setlength{\tabcolsep}{2.6pt}
\renewcommand{\arraystretch}{1.03}
\resizebox{\columnwidth}{!}{%
\begin{tabular}{@{}lcccc@{}}
\toprule
& \multicolumn{4}{c}{\textbf{Cultures}} \\
\midrule
\textbf{Metric} & \textbf{Indian} & \textbf{Japanese} & \textbf{Italian} & \textbf{Turkish} \\
\midrule
\multicolumn{5}{c}{\textbf{($\S$) ALaDiT OneHot}} \\
\midrule
FGD $\downarrow$ & $3.68 \pm 0.61$ & {\boldmath$2.44 \pm 0.26$}$^{\ddagger,\dagger,\P}$ & $2.68 \pm 0.50$ & $3.91 \pm 0.69$$^{\ast}$ \\
CE F1(\%) $\uparrow$ & $50.94 \pm 1.55$$^{\ast}$ & $43.90 \pm 2.23$$^{\ast,\circ}$ & $37.51 \pm 2.21$$^{\ast}$ & $42.57 \pm 1.13$$^{\ddagger}$ \\
BAS(\%) $\uparrow$ & $22.28 \pm 0.27$ & $22.37 \pm 0.42$ & $22.58 \pm 0.57$ & {\boldmath$22.85 \pm 0.28$} \\
SRGR(\%) $\uparrow$ & $70.10 \pm 0.45$$^{\ddagger}$ & $67.07 \pm 0.87$ & $71.69 \pm 0.76$$^{\ast}$ & $63.02 \pm 0.64$$^{\ast}$ \\
Diversity $\uparrow$ & {\boldmath$107.18 \pm 2.70$}$^{\ddagger,\dagger}$ & $114.62 \pm 2.35$$^{\ddagger,\dagger,\circ}$ & $112.64 \pm 1.23$$^{\ddagger,\dagger,\circ}$ & $112.45 \pm 1.45$$^{\ddagger,\dagger}$ \\
\midrule
\multicolumn{5}{c}{\textbf{($\P$) ALaDiT NoDG}} \\
\midrule
FGD $\downarrow$ & $2.80 \pm 0.52$$^{\ddagger,\S}$ & $2.96 \pm 0.29$ & $2.79 \pm 0.48$ & $4.00 \pm 0.59$$^{\ast}$ \\
CE F1(\%) $\uparrow$ & $49.85 \pm 1.34$ & $42.93 \pm 2.02$ & $36.58 \pm 2.20$ & $43.35 \pm 1.73$$^{\ddagger}$ \\
BAS(\%) $\uparrow$ & $22.30 \pm 0.26$ & $22.53 \pm 0.38$ & $22.53 \pm 0.44$ & $22.74 \pm 0.46$ \\
SRGR(\%) $\uparrow$ & $70.41 \pm 0.38$$^{\ddagger,\S}$ & $67.13 \pm 0.75$ & $71.69 \pm 0.74$$^{\ast}$ & $63.06 \pm 0.54$$^{\ast}$ \\
Diversity $\uparrow$ & $106.39 \pm 2.21$$^{\dagger}$ & {\boldmath$114.67 \pm 2.63$}$^{\ddagger,\dagger,\circ}$ & $112.22 \pm 0.95$$^{\ddagger,\dagger,\circ}$ & $113.09 \pm 1.25$$^{\ddagger,\dagger}$ \\
\midrule
\multicolumn{5}{c}{\textbf{($\circ$) ALaDiT NoAlign}} \\
\midrule
FGD $\downarrow$ & $2.87 \pm 0.47$$^{\ddagger,\S}$ & $2.45 \pm 0.23$$^{\ddagger,\dagger,\P}$ & {\boldmath$2.09 \pm 0.54$}$^{\ast,\S,\P}$ & $3.12 \pm 0.47$$^{\ddagger,\ast,\S,\P}$ \\
CE F1(\%) $\uparrow$ & $51.01 \pm 1.21$$^{\ast}$ & $42.08 \pm 2.09$ & $36.56 \pm 1.85$ & {\boldmath$43.84 \pm 1.24$}$^{\ddagger,\dagger,\S}$ \\
BAS(\%) $\uparrow$ & $22.34 \pm 0.25$ & $22.50 \pm 0.29$ & $22.76 \pm 0.26$ & $22.82 \pm 0.35$ \\
SRGR(\%) $\uparrow$ & {\boldmath$70.75 \pm 0.41$}$^{\ddagger,\ast,\S,\P}$ & $67.63 \pm 0.77$$^{\ast,\S,\P}$ & {\boldmath$72.36 \pm 0.61$}$^{\ddagger,\ast,\S,\P}$ & $63.35 \pm 0.57$$^{\ast,\S}$ \\
Diversity $\uparrow$ & $106.28 \pm 2.34$$^{\dagger}$ & $113.66 \pm 2.52$$^{\ddagger}$ & $111.60 \pm 1.29$ & $112.50 \pm 1.30$$^{\ddagger,\dagger}$ \\
\midrule
\multicolumn{5}{c}{\textbf{($\ddagger$) ALaDiT NC}} \\
\midrule
FGD $\downarrow$ & $3.96 \pm 0.65$ & $3.45 \pm 0.61$ & $2.31 \pm 0.38$$^{\ast,\P}$ & $3.94 \pm 0.49$ \\
CE F1(\%) $\uparrow$ & $51.54 \pm 1.25$$^{\ast,\P}$ & $44.17 \pm 2.45$$^{\ast,\circ}$ & {\boldmath$37.92 \pm 1.79$}$^{\ast}$ & $40.00 \pm 1.48$ \\
BAS(\%) $\uparrow$ & $22.37 \pm 0.28$ & $22.36 \pm 0.39$ & $22.67 \pm 0.55$ & $22.69 \pm 0.26$ \\
SRGR(\%) $\uparrow$ & $69.39 \pm 0.33$ & {\boldmath$67.66 \pm 0.76$}$^{\ast,\S,\P}$ & $71.78 \pm 0.85$$^{\ast}$ & {\boldmath$63.69 \pm 0.67$}$^{\ast,\dagger,\S,\P,\circ}$ \\
Diversity $\uparrow$ & $105.76 \pm 2.36$ & $111.43 \pm 2.59$ & $111.11 \pm 1.05$ & $110.99 \pm 1.06$ \\
\midrule
\multicolumn{5}{c}{\textbf{($\ast$) ALaDiT ADV}} \\
\midrule
FGD $\downarrow$ & {\boldmath$2.70 \pm 0.51$}$^{\ddagger,\S,\circ}$ & $2.46 \pm 0.34$$^{\ddagger,\dagger,\P}$ & $2.81 \pm 0.41$ & $4.27 \pm 0.60$ \\
CE F1(\%) $\uparrow$ & $49.38 \pm 0.98$ & $42.25 \pm 1.65$ & $35.88 \pm 1.58$ & $43.32 \pm 1.28$$^{\ddagger}$ \\
BAS(\%) $\uparrow$ & $22.37 \pm 0.37$ & $22.22 \pm 0.29$ & $22.41 \pm 0.41$ & $22.75 \pm 0.37$ \\
SRGR(\%) $\uparrow$ & $70.36 \pm 0.37$$^{\ddagger}$ & $67.21 \pm 0.85$ & $71.20 \pm 0.58$ & $62.64 \pm 0.63$ \\
Diversity $\uparrow$ & $106.78 \pm 2.58$$^{\dagger}$ & $114.20 \pm 2.79$$^{\ddagger,\dagger}$ & {\boldmath$112.82 \pm 1.30$}$^{\ddagger,\dagger,\P,\circ}$ & {\boldmath$113.49 \pm 1.49$}$^{\ddagger,\dagger,\S,\circ}$ \\
\midrule
\multicolumn{5}{c}{\textbf{($\dagger$) ALaDiT FI}} \\
\midrule
FGD $\downarrow$ & $2.71 \pm 0.47$$^{\ddagger,\S}$ & $3.03 \pm 0.27$$^{\ddagger}$ & $2.34 \pm 0.60$$^{\ast,\P}$ & {\boldmath$2.89 \pm 0.43$}$^{\ddagger,\ast,\S,\P,\circ}$ \\
CE F1(\%) $\uparrow$ & {\boldmath$52.44 \pm 1.42$}$^{\ddagger,\ast,\S,\P,\circ}$ & {\boldmath$45.83 \pm 1.57$}$^{\ddagger,\ast,\S,\P,\circ}$ & $37.36 \pm 1.57$$^{\ast}$ & $42.79 \pm 1.55$$^{\ddagger}$ \\
BAS(\%) $\uparrow$ & {\boldmath$22.44 \pm 0.34$} & {\boldmath$22.57 \pm 0.35$} & {\boldmath$22.93 \pm 0.56$}$^{\ast,\P}$ & $22.74 \pm 0.35$ \\
SRGR(\%) $\uparrow$ & $70.66 \pm 0.41$$^{\ddagger,\ast,\S}$ & $67.52 \pm 0.76$$^{\ast,\S,\P}$ & $72.20 \pm 0.82$$^{\ast,\S,\P}$ & $63.34 \pm 0.57$$^{\ast,\S,\P}$ \\
Diversity $\uparrow$ & $105.19 \pm 2.42$ & $112.81 \pm 2.57$ & $111.32 \pm 0.98$ & $111.19 \pm 0.90$ \\
\bottomrule
\end{tabular}%
}
\caption{Per-culture ALaDiT ablations, reported as mean $\pm$ std over 10 matched test runs. Bold marks the best value within each culture and metric across the ALaDiT block. Superscripts denote values significantly better than the marked row under paired two-sided $t$-tests ($p<0.01$). Per-culture CE F1 is the one-vs-rest F1 for the corresponding culture class; BAS and SRGR are reported as percentages.}
\label{tab:per_culture_metrics}
\end{table}

\begin{table}[htbp]
\centering
\scriptsize
\setlength{\tabcolsep}{6pt}
\renewcommand{\arraystretch}{1}
\resizebox{0.98\columnwidth}{!}{%
\begin{tabular}{l@{\hspace{0.6cm}} *{4}{c}}
\toprule
& \multicolumn{4}{c}{\textbf{Cultures}} \\
\midrule
\textbf{Questions} & \textbf{Indian} & \textbf{Japanese} & \textbf{Italian} & \textbf{Turkish} \\
\midrule
\multicolumn{5}{c}{\textbf{ALaDiT FI}} \\
\midrule
Coherence with Speech $\uparrow$ & {\boldmath$5.33 \pm 1.75$} & {\boldmath$6.35 \pm 1.93$}$^{\ddagger}$ & $5.90 \pm 2.54$ & $6.00 \pm 2.17$ \\
Appropriateness $\uparrow$ & {\boldmath$5.58 \pm 1.50$} & {\boldmath$6.67 \pm 1.84$}$^{\ddagger}$ & {\boldmath$6.05 \pm 2.29$} & $6.17 \pm 1.42$ \\
Fluency $\uparrow$ & {\boldmath$5.83 \pm 1.64$} & {\boldmath$6.22 \pm 1.54$} & $6.30 \pm 2.17$ & $6.38 \pm 1.48$ \\
Timing $\uparrow$ & {\boldmath$5.38 \pm 1.59$} & {\boldmath$6.22 \pm 1.77$} & {\boldmath$6.22 \pm 2.05$} & $6.05 \pm 1.41$ \\
Amount of Gesticulation $\uparrow$ & {\boldmath$5.60 \pm 1.62$} & {\boldmath$6.20 \pm 1.85$}$^{\ddagger}$ & $5.70 \pm 2.11$ & $5.97 \pm 1.82$ \\
Naturalness $\uparrow$ & {\boldmath$6.22 \pm 1.69$}$^{\ast}$ & {\boldmath$6.55 \pm 1.45$}$^{\ddagger}$ & $6.10 \pm 2.26$ & $6.12 \pm 1.80$ \\
Cultural Match $\uparrow$ & {\boldmath$5.83 \pm 1.50$} & {\boldmath$6.45 \pm 2.32$}$^{\ddagger}$ & {\boldmath$6.15 \pm 2.87$} & $6.20 \pm 1.61$ \\
\midrule
\multicolumn{5}{c}{\textbf{ALaDiT ADV}} \\
\midrule
Coherence with Speech $\uparrow$ & $5.30 \pm 1.48$ & $5.38 \pm 1.36$$^{\dagger}$ & {\boldmath$6.00 \pm 2.25$} & $5.72 \pm 1.59$ \\
Appropriateness $\uparrow$ & $4.95 \pm 1.99$ & $5.58 \pm 1.52$$^{\dagger}$ & $5.92 \pm 1.93$ & $5.62 \pm 1.31$$^{\ast}$ \\
Fluency $\uparrow$ & $5.55 \pm 1.98$ & $5.60 \pm 1.52$ & {\boldmath$6.40 \pm 1.45$} & $6.12 \pm 1.57$ \\
Timing $\uparrow$ & $4.90 \pm 1.80$ & $5.62 \pm 1.40$ & $6.08 \pm 1.60$ & $6.17 \pm 1.48$ \\
Amount of Gesticulation $\uparrow$ & $5.05 \pm 2.28$ & $5.00 \pm 1.95$$^{\dagger,\ast}$ & $5.53 \pm 2.22$ & $5.78 \pm 2.17$ \\
Naturalness $\uparrow$ & $5.67 \pm 1.82$ & $5.70 \pm 1.45$$^{\dagger}$ & {\boldmath$6.25 \pm 1.94$} & $5.83 \pm 1.42$$^{\ast}$ \\
Cultural Match $\uparrow$ & $5.20 \pm 1.41$ & $5.42 \pm 1.63$$^{\dagger}$ & $5.95 \pm 2.21$ & $6.03 \pm 1.59$ \\
\midrule
\multicolumn{5}{c}{\textbf{ALaDiT NC}} \\
\midrule
Coherence with Speech $\uparrow$ & $5.17 \pm 2.20$ & $5.70 \pm 1.73$ & $5.75 \pm 1.96$ & {\boldmath$6.20 \pm 1.94$} \\
Appropriateness $\uparrow$ & $5.12 \pm 2.16$ & $5.95 \pm 1.42$ & $5.90 \pm 1.54$ & {\boldmath$6.50 \pm 1.69$}$^{\ddagger}$ \\
Fluency $\uparrow$ & $5.42 \pm 2.49$ & {\boldmath$6.22 \pm 1.80$} & $5.58 \pm 1.30$ & {\boldmath$6.42 \pm 2.01$} \\
Timing $\uparrow$ & $4.80 \pm 2.34$ & $6.05 \pm 1.96$ & $6.17 \pm 1.65$ & {\boldmath$6.50 \pm 1.85$} \\
Amount of Gesticulation $\uparrow$ & $5.08 \pm 2.61$ & $6.00 \pm 2.24$$^{\ddagger}$ & {\boldmath$5.72 \pm 1.56$} & {\boldmath$6.58 \pm 2.01$} \\
Naturalness $\uparrow$ & $4.97 \pm 2.16$$^{\dagger}$ & $6.28 \pm 1.79$ & $5.83 \pm 1.64$ & {\boldmath$6.83 \pm 2.14$}$^{\ddagger}$ \\
Cultural Match $\uparrow$ & $5.15 \pm 2.13$ & $5.83 \pm 2.31$ & $5.90 \pm 1.86$ & {\boldmath$6.38 \pm 2.15$} \\
\midrule
\multicolumn{5}{c}{\textbf{Real}} \\
\midrule
Coherence with Speech $\uparrow$ & $6.88 \pm 2.13$$^{\S}$ & $6.72 \pm 2.04$$^{\S}$ & $7.45 \pm 1.97$$^{\S}$ & $6.75 \pm 1.98$$^{\S}$ \\
Appropriateness $\uparrow$ & $6.83 \pm 1.64$$^{\S}$ & $6.58 \pm 2.03$ & $7.72 \pm 1.80$$^{\S}$ & $6.88 \pm 1.53$$^{\S}$ \\
Fluency $\uparrow$ & $6.85 \pm 1.74$$^{\S}$ & $7.08 \pm 1.64$$^{\S}$ & $7.72 \pm 1.85$$^{\S}$ & $6.95 \pm 1.74$$^{\S}$ \\
Timing $\uparrow$ & $6.97 \pm 1.35$$^{\S}$ & $6.70 \pm 1.73$$^{\S}$ & $7.80 \pm 1.41$$^{\S}$ & $6.97 \pm 1.76$$^{\S}$ \\
Amount of Gesticulation $\uparrow$ & $6.15 \pm 2.32$$^{\S}$ & $6.92 \pm 2.18$$^{\S}$ & $7.50 \pm 1.99$$^{\S}$ & $7.03 \pm 1.85$$^{\S}$ \\
Naturalness $\uparrow$ & $6.88 \pm 1.68$$^{\S}$ & $7.15 \pm 1.73$$^{\S}$ & $7.47 \pm 1.84$$^{\S}$ & $6.72 \pm 1.60$ \\
Cultural Match $\uparrow$ & $6.72 \pm 2.04$$^{\S}$ & $6.28 \pm 1.96$ & $7.88 \pm 1.78$$^{\S}$ & $7.05 \pm 2.08$$^{\S}$ \\
\bottomrule
\end{tabular}%
}
\caption{Per-culture user-study ratings for the three generated ALaDiT models (FI, ADV, and NC) plus the Real reference motion. Values are participant-level mean $\pm$ std over 20 participants; repeated trials from the same participant for the same question, culture, and model are averaged before aggregation. Within each culture and question, the best generated model is bolded. Real values are not bolded; $^\S$ marks cells where Real exceeds the best generated model for the same culture and question. Superscripts on generated-model cells mark paired t-test significance at $p<0.05$: $^\dagger$ vs ALaDiT FI, $^\ddagger$ vs ALaDiT ADV, and $^\ast$ vs ALaDiT NC. All scores use the original 0--10 user-study scale.}
\label{tab:user_study_per_culture_generated}
\end{table}

We further analyze the results separately for each culture to better interpret the findings reported in the main paper. In the main paper, we compare the three principal ALaDiT variants, ALaDiT NC, ALaDiT ADV, and ALaDiT FI, together with three ablations: OneHot, NoDG, and NoAlign. Results show that FI provides the strongest overall trade-off, achieving the best FGD, CE F1, and BAS, while remaining competitive on SRGR and Diversity. NoAlign obtains the highest SRGR, whereas OneHot and ADV mainly increase Diversity. The user study, conducted with $N=20$ participants, also supports this interpretation: FI obtains the highest average score among generated models, is significantly preferred over ADV overall, and is significantly preferred over NC on Cultural Match. The per-culture results in Table~\ref{tab:per_culture_metrics} and Table~\ref{tab:user_study_per_culture_generated} show how these trends vary across cultural groups.

Overall, the per-culture objective results are consistent with the main conclusion, but they also show that the effect of cultural conditioning is not uniform across cultures or metrics. FI remains the most balanced variant: it gives the strongest results overall, with the best CE F1 for Indian and Japanese samples, the best BAS for Indian, Japanese, and Italian samples, and the best FGD for Turkish samples. However, some isolated metrics are optimized by other variants for specific cultures. NoAlign reaches the best SRGR for Indian and Italian samples and the best FGD for Italian samples, showing that removing the alignment losses can sometimes improve specific per-culture metrics, although its aggregate FGD and CE F1 remain inferior to FI. OneHot and NoDG occasionally obtain the highest Diversity, while ADV remains competitive mainly in Diversity and in some per-culture FGD cases. NC is also not uniformly worse at the per-culture level, since it remains best for some isolated cases, such as Italian CE F1 and Japanese/Turkish SRGR.  These results therefore support the interpretation that FI gives the most reliable gains overall, while per-culture behavior remains heterogeneous.

The per-culture user-study results follow the same pattern of overall improvement with substantial cultural heterogeneity. FI obtains the strongest ratings for Indian samples across all questions, and for Japanese samples, it is best or tied for all questions, with several significant advantages over ADV. For Italian samples, FI is strongest on Appropriateness, Timing, and Cultural Match, while ADV and NC lead some other perceptual questions.Turkish samples are the main exception: NC obtains the highest mean score on most questions, although the differences with FI are not statistically significant. This explains why the aggregate user-study result favors FI while the per-culture table still contains mixed local rankings. This pattern should be interpreted cautiously, since participants may not be equally familiar with all evaluated cultural styles.

Including the Real reference clarifies the absolute scale of the perceptual results. Real motion usually remains above the best generated model, especially for Italian samples and for several Indian, Japanese, and Turkish questions, confirming that generated gestures are still perceptually distinguishable from ground-truth motion. At the same time, the gaps are not equally large for every culture and question, suggesting that generated motion can approach real-motion ratings in some settings.

Overall, the supplementary results reinforce the main-paper conclusion: Fishr based cultural embeddings provide the most reliable trade-off between realism, cultural consistency, and perceptual quality, while the per-culture analysis shows that these gains are not homogeneous across all cultural groups.



\end{document}